\documentclass[runningheads]{llncs}

\usepackage{eccv}

\usepackage{scrextend}%

\usepackage{eccvabbrv}
\usepackage{xspace}
\usepackage{graphicx}
\usepackage{booktabs}
\usepackage[export]{adjustbox}

\usepackage[accsupp]{axessibility}  %

\usepackage[hypertexnames=false]{hyperref}

\usepackage{tabularx} %
\usepackage[labelsep=period,font=small]{caption} %
\usepackage{tikz} %
\usepackage{tikzfill} %
\usepackage{pgfplots} %
\usepackage{pgfplotstable} %
\usepackage{standalone} %
\usepackage{multirow,makecell} %
\usepackage{tabularx}
\usepackage{booktabs}
\usepackage{siunitx}

\newcolumntype{Y}{>{\centering\arraybackslash}X} %
\newcolumntype{P}[1]{>{\centering\arraybackslash}p{#1}} %
\def\checkmark{\tikz\fill[scale=0.4](0,.35) -- (.25,0) -- (1,.7) -- (.25,.15) -- cycle;} %
\def\crossmark{\tikz{\draw[line width=1pt](0,0.2) -- (0.2,0);\draw[line width=1pt](0,0) -- (0.2,0.2);}} %

\usetikzlibrary{spy} %
\usetikzlibrary{arrows.meta} %
\usepgfplotslibrary{polar} %
\usepgfplotslibrary{statistics}

\pgfplotsset{compat=newest} %

\usetikzlibrary{positioning}
\usetikzlibrary{calc}
\usepgfplotslibrary{fillbetween}
\usetikzlibrary{patterns}
\usepgfplotslibrary{colorbrewer}
\usetikzlibrary{positioning}
\usetikzlibrary{shapes.geometric, arrows, spath3, intersections, calc}
\usetikzlibrary{fit}
\usepgfplotslibrary{groupplots}
\usetikzlibrary{decorations.markings}

\definecolor{3787CF}{RGB}{55,135,207}
\definecolor{F1C232}{RGB}{241,194,50}
\definecolor{9D4761}{RGB}{157,71,97}
\definecolor{619D47}{RGB}{97,157,71}

\pgfplotscreateplotcyclelist{custompalette}{
    {3787CF!90!black,line width=1.5pt},
    {F1C232!90!black,line width=1.5pt},
    {9D4761!90!black,line width=1.5pt},
    {619D47!90!black,line width=1.5pt}
}

\pgfplotscreateplotcyclelist{custompalettethin}{
    {3787CF!90!black,line width=1.0pt},
    {F1C232!90!black,line width=1.0pt},
    {9D4761!90!black,line width=1.0pt},
    {619D47!90!black,line width=1.0pt}
}

\tikzstyle{arrow} = [->,>=stealth]

\usepackage{listings}

\usepackage{comment}

\NewDocumentCommand{\headerot}{O{45} O{1em} m}{\makebox[#2][l]{\rotatebox{#1}{#3}}}%

\usepackage{stmaryrd}
\usepackage{trimclip}

\makeatletter
\DeclareRobustCommand{\shortto}{%
  \mathrel{\mathpalette\short@to\relax}%
}

\DeclareRobustCommand{\veryshortto}{%
  \mathrel{\mathpalette\veryshort@to\relax}%
}

\newcommand{\short@to}[2]{%
  \mkern2mu
  \clipbox{{.3\width} 0 0 0}{$\m@th#1\vphantom{+}{\shortrightarrow}$}%
  }

\newcommand{\veryshort@to}[2]{%
  \mkern2mu
  \clipbox{{.2\width} 0 0 0}{$\m@th#1\vphantom{+}{\shortrightarrow}$}%
  }
\makeatother

\usepackage{soul}

\usepackage{graphicx}%
\usepackage{multirow}%
\usepackage{amsmath,amssymb,amsfonts}%
\usepackage{mathrsfs}%
\usepackage[title]{appendix}%
\usepackage{xcolor}%
\usepackage{textcomp}%
\usepackage{manyfoot}%
\usepackage{booktabs}%
\usepackage{algorithm}%
\usepackage{algorithmicx}%
\usepackage{algpseudocode}%
\usepackage{listings}%
\usepackage{calc}
\usepackage{fontawesome7}
\usepackage{xspace}

\usepackage{siunitx}
\usepackage{nicefrac}

\usepackage{etoolbox}           %

\usepackage[normalem]{ulem}

\robustify\uline

\usepackage[capitalize]{cleveref}
\crefname{section}{Sec.}{Secs.}
\Crefname{section}{Section}{Sections}
\Crefname{table}{Table}{Tables}
\crefname{table}{Tab.}{Tabs.}

\usepackage{orcidlink}
\usepackage[nohyperlinks,printonlyused]{acronym}

\makeatletter
\DeclareRobustCommand\onedot{\futurelet\@let@token\@onedot}
\def\@onedot{\ifx\@let@token.\else.\null\fi\xspace}

\def\eg{\emph{e.g}\onedot} 
\def\ie{\emph{i.e}\onedot} 
\def\cf{\emph{cf}\onedot} 
 
\def\wrt{w.r.t\onedot} 
 
\def\etal{\emph{et al}\onedot}

\newcommand*{\citelinktext}[2]{%
  \nocite{#1}%
  \hyper@@link[cite]{}{cite.#1}{#2}%
}

\definecolor{LIGHTPINK}{RGB}{237,157,202}
\definecolor{LIGHTRED}{RGB}{210,121,121}
\definecolor{LIGHTORANGE}{RGB}{230,170,50}
\definecolor{LIGHTGOLD}{RGB}{210,194,121}
\definecolor{LIGHTGREEN}{RGB}{121,210,121}
\definecolor{LIGHTAQUA}{RGB}{121,206,210}
\definecolor{LIGHTBLUE}{RGB}{121,124,210}
\definecolor{LIGHTPURPLE}{RGB}{153,102,255}
\definecolor{RED}{RGB}{178,34,34}
\definecolor{GRAY}{RGB}{166,166,166}
\definecolor{WHITE}{RGB}{255,255,255}

\DeclareSIUnit{\fps}{ \translate{FPS} }

\let\titleold\title
\renewcommand{\title}[1]{\titleold{#1}\newcommand{\thetitle}{#1}}
\def\maketitlesupplementary
   {
   \newpage
        {\centering
        \Large
        \textbf{\thetitle}\\
        \vspace{0.5em}Supplementary Material \\
        \vspace{1.0em}
        }
    }

\makeatother

\newcommand{\wiflow}{WiFlow\xspace}

\def\colwidth{122.0mm}

\newcommand{\ours}{{\wiflow}$_{\text{Simple}}$\xspace}
\newcommand{\ourm}{{\wiflow}$_{\text{RoI}}$\xspace}
\newcommand{\ourl}{{\wiflow}$_{\text{Combo}}$\xspace}
\newcommand{\sideview}{\textit{sideview}\xspace}
\newcommand{\birdview}{\textit{birdview}\xspace}
\newcommand{\birdviewplus}{\textit{birdview$_+$}\xspace}
\newcommand{\pseudo}{Pseudo$_{\text{GT}}$\xspace}
\newcommand{\noise}{\text{off-area}\xspace}
\newcommand{\void}{\text{void}\xspace}

\newcommand{\wifi}{WiFi\xspace}

\definecolor{context}{HTML}{d95f02}
\definecolor{feature}{HTML}{1b9e77}
\definecolor{refine}{HTML}{7570b3}

\definecolor{flow}{HTML}{66c2a5}
\definecolor{mask}{HTML}{8da0cb}

\definecolor{simple}{HTML}{66c2a5}
\definecolor{roi}{HTML}{8da0cb}
\definecolor{combo}{HTML}{fc8d62}
\definecolor{zero}{HTML}{e78ac3}

\usepackage{placeins} %

\usepackage{xprintlen}

\usepackage{fancyhdr}

\fancypagestyle{acceptedfooter}{
    \fancyhf{} %

    \fancyfoot[L]{
        \vspace{1.5em}
        \scriptsize
        \framebox{\parbox{.98\linewidth}{To appear in Proceedings of the \emph{19th European Conference on Computer Vision (ECCV)}, 2026. The final publication will be available through Springer.}}
    }
}

\begin{document}

\title{\wiflow: Estimating Optical Flow using \wifi Channel State Information} 

\titlerunning{\wiflow: Estimating Optical Flow using \wifi CSI}

\author{
Thomas Weigel\inst{1}\orcidlink{0009-0002-0446-2643} \and
Simon Kiefhaber\inst{1,3}\orcidlink{0009-0000-7659-6435} \and
Fabian Portner\inst{2}\orcidlink{0009-0002-1984-9533} \and
Matthias Hollick\inst{1}\orcidlink{0000-0002-9163-5989} \and
Simone Schaub-Meyer\inst{1,3}\orcidlink{0000-0001-8644-1074}
}

\authorrunning{Weigel et al.}

\institute{
    Technical University of Darmstadt, Germany\\
    \email{weigel-thomas@outlook.de, \{name.surname\}@tu-darmstadt.de}
    \and
    Technische Universiteit Delft, The Netherlands\\
    \email{fportner@tudelft.nl}
    \and
    Hessian Center for AI (hessian.AI), Germany
}

\maketitle
\thispagestyle{acceptedfooter}
\begin{figure}
    \centering
    \includegraphics[width=0.95\linewidth]{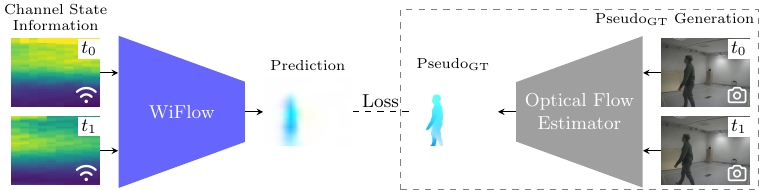}
    \caption{\textbf{\wiflow.} Can optical flow be estimated solely from \wifi information? Overview of our framework to estimate optical flow from channel state information extracted from \wifi signals. Pseudo ground truth to train \wiflow is computed from video frames at two corresponding timestamps.
    }
    \label{fig:overview}
\end{figure}

\begin{abstract}
Knowing where and how fast objects are moving within a scene is important across various domains. Usually, cameras are used to capture the data necessary for this task, but adding cameras often raises privacy concerns, and the quality of captured frames is heavily influenced by lighting conditions. In this work, we explore using WiFi channel state information (CSI) instead of camera frames for optical flow estimation. We propose WiFlow, a CSI based flow estimator, a preprocessor evaluation for CSI, and three model architectures that offer different trade-offs between accuracy and complexity. Further, we create the first dataset for training and evaluating CSI-based optical flow estimators, and our experiments provide insights into key design elements for this task. Code and data are available at \url{https://visinf.github.io/wiflow}.

\end{abstract}

\section{Introduction}
\label{sec:intro}

Motion is a fundamental cue for understanding dynamic scenes. In computer vision, a standard way to capture temporal change is optical flow, which estimates the apparent motion between two consecutive frames. Optical flow supports a wide range of downstream tasks, such as action recognition~\cite{Zhu:2018:HTS}, object tracking~\cite{FOLT:2023}, robotic navigation and control~\cite{6GObjectTrackRobo:2023,Drone2024,PolarMendoza:2025}, inpainting~\cite{Gao:2020:FGV,Ke:2021:OAV,Xu:2019:DFG,Zhou:2023:PIP}, video frame interpolation~\cite{SoftmaxSplatting:2020,Dong:2023:VFI,Sim:2021:XVF,Zhang:2023:EMA,Niklaus:2018:CAS}, and unsupervised segmentation~\cite{Hahn:2025:SCU,Choudhury:2022:GWM,Safadoust:2023:MOD}. %
In practice, optical flow is almost always computed from camera images, which makes deployments in certain settings difficult. Cameras record detailed visual scenes, expose people's privacy and fail in the dark without additional hardware, as we illustrate in the supplemental \cref{supp:brightness}.

\wifi sensing offers a different way to observe motion, especially indoors. \wifi devices continuously estimate channel state information (CSI), which captures how signals change as they propagate from transmitter to receiver. Indoors, this propagation is determined by the influence of static objects, like walls and furniture, but also moving subjects, like people. When something moves, the propagation changes, and this is reflected in the CSI measurements. This relation makes CSI a natural input for device-free sensing tasks, requiring no additional hardware beyond what is already used for wireless communication.

In this work, we ask: Can optical flow be estimated solely from \wifi information? Prior CSI-based sensing work shows that \wifi signals carry elaborate motion information, enabling localization~\cite{widir:2016,wispeed:2018}, gesture and action recognition~\cite{witraj:2023,WiDar3,csi-gesture}, and human pose estimation~\cite{mmfi2023,person-in-wifi:2019,wipe:2025,hpe-li-pose:2024}. These CSI-based methods directly produce a task-specific output, such as a class, a location, or a pose. Here, we push further and instead try to recover optical flow from CSI directly, capturing the motion pattern itself rather than a single task output.

Estimating flow from \wifi is especially attractive when cameras are undesirable or unreliable. \wifi sensing works in the dark and can build on existing wireless infrastructure, while avoiding the collection of camera frames. If dense flow from CSI is feasible, it can serve as a general motion input for privacy-sensitive indoor applications such as home security, fall detection, gesture control, or crowd monitoring.

In this work, we demonstrate the feasibility of extracting general motion information from CSI measurements in a fixed indoor setting. Our main contributions are 
\emph{(i)} experimental evidence that optical flow can be recovered from CSI with generalization across unseen subjects, 
\emph{(ii)} three distinct architectures for CSI-based optical flow estimation, offering varying trade-offs between accuracy and computational complexity, and
\emph{(iii)} a novel dataset providing optical flow supervision (pseudo ground truth) alongside synchronized CSI for training and evaluation.

\section{Related Work}
\subsubsection{Optical Flow} is an estimate of the apparent pixel-wise motion between consecutive image frames. Methods for computing it range from classical formulations~\cite{Horn:1981:DOF, Black:1996:REM,lucaskanade} to deep learning approaches, which are typically built on either convolutional or transformer-based architectures.

On the convolutional side, FlowNet~\cite{Dosovitskiy:2015:FN} introduced learned optical flow estimation, and later work such as PWC-Net~\cite{Sun:2018:PWC,Sun:2020:MMS}, FlowNet2~\cite{Ilg:2017:FN2}, and related methods~\cite{Hur:2019:IRR,spynet:2017} refined the same core pipeline. Many of these models reduce the cost of dense matching with coarse-to-fine estimation and restricted per-pixel search ranges, which can limit the magnitude of motion they capture. RAFT~\cite{RAFT} avoided this fixed-range design by using global correlation with iterative refinement, and subsequent RAFT-style methods~\cite{SEARAFT,recover:2025,waft:2026,gmflow,gma:2021} improved robustness in challenging cases such as occlusions~\cite{gma:2021}. Transformer-based alternatives are also emerging~\cite{flowformer++2023,Memflow2024}, though recent comparisons still report convolutional methods as strong in both accuracy and computational cost~\cite{SEARAFT,recover:2025,DPFlow2025}. All of these methods infer motion from video by tracking correspondence across frames. Our goal is to bypass cameras altogether and extract comparable flow estimates from \wifi channel state information alone.

\subsubsection{\wifi Sensing.}

Several works reconstruct camera-like observations from channel state information (CSI), including RGB frames~\cite{wi2vi:2020,mwi2vi:2022,csi2image:2021} and dense geometry via depth images~\cite{csi2depth:2025}. While some pipelines also produce person-shaped masks or saliency maps~\cite{wi2vi:2020,mwi2vi:2022} that can look motion-like, their objective is still frame or depth synthesis rather than learning pixel correspondences \textit{across time}. In parallel, CSI-to-human methods estimate human-centric outputs such as person masks/segmentation~\cite{person-in-wifi:2019}, 3D pose or skeleton tracking~\cite{wipose:2020,gopose:2022,winect:2022,person-in-wifi-3d:2024,hpe-li-pose:2024,wipe:2025,csipose:2025,metafi:2022}, and full 3D mesh construction~\cite{multimesh:2024}. Finally, completion approaches fuse \wifi CSI with vision for RGB recovery (inpainting or occlusion removal) and require an image input for inference~\cite{rfinpainter:2022,csiinpainter:2025}.

Separately, CSI has also been used to infer \emph{motion in physical space}. Representative systems track device-free motion trajectories~\cite{widar:2017,widar2:2018,jindirection:2020,witraj:2023}, estimate specific attributes such as walking direction~\cite{widir:2016} or speed and acceleration~\cite{wispeed:2018,wivelo:2024}, and perform device-free localization~\cite{lifs:2016}. These outputs are expressed in metric coordinates and are not designed to produce a dense image-plane motion representation.

Overall, prior work either reconstructs spatial content (RGB/depth/meshes) or predicts motion in metric space (\eg, trajectories). In contrast, we learn \emph{CSI-to-optical-flow} to predict dense image-plane motion over the sensed area: video is only used to generate pseudo-flow supervision during training, at inference, we output optical flow from CSI alone.

\subsubsection{Preprocessing.}
\label{sec:preprocessing}

Raw CSI is hard to learn from. As a channel estimate, it contains both environmental effects and measurement/hardware artifacts that are unrelated to the scene. This is why \wifi sensing pipelines usually apply preprocessing to reduce these artifacts and make the representation easier to learn from, even when the downstream model is a neural network. In the simplest case, this is ignored, and CSI amplitudes and phases are being used without any processing~\cite{deepfi:2017,metafi:2022}. Some works apply time--frequency transforms to expose motion-related spectral structure. \textit{Fourier} transformations (\eg, STFT-style processing) have been used to extract micro-Doppler signatures imprinted on CSI evolution in time~\cite{wisee:2013,carm:2015,widance:2017}. Alternatively, wavelet transforms have been used both to derive multi-scale (scale--frequency) features for learning~\cite{wistep:2018,freesense:2018} and to denoise CSI in the wavelet domain before subsequent processing~\cite{wifree:2018,crowdcount:2018}. Further, Principal Component Analysis (PCA) is often used to either reduce dimensionality ~\cite{wikey:2015}, or as a filtering step that suppresses a dominant common component when it mainly captures nuisance variation~\cite{carm:2015}. Some phase offsets are common to all antennas on a device, \eg, due to shared hardware. To mitigate these, multi-antenna normalization is often used, either through antenna conjugation~\cite{widance:2017,indotrack:2017} or quotients~\cite{witraj:2023}. Finally, to combat noise, smoothing filters such as \textit{Savitzky--Golay} are frequently applied to reduce high-frequency noise while preserving local structure~\cite{widir:2016,witraj:2023,pulsefi:2025}.

\section{\wifi Sensing}
While \wifi packets are primarily designed to carry communication data, they simultaneously carry a fingerprint of the environment. 
The transmitter encodes data as a symbol $x$, which undergoes complex transformations as it propagates through the physical environment. Consequently, the receiver does not observe $x$ directly, but rather a distorted version shaped by the surrounding environment. Conceptually,
\begin{equation}
    y = Hx + n,
\end{equation}
where $y$ is what the receiver measures, $n$ denotes noise, and $H$ is the wireless channel. The channel $H$ represents how the current environment transforms the transmitted information $x$ as it travels from transmitter to receiver. When the environment changes, propagation changes, and so does $H$. For communication, $H$ is a nuisance: the receiver needs the information in $x$, not the transformed observation $y$. This is why every \wifi packet begins with a short, fixed, known sequence. The receiver compares the expected and received symbols to estimate $H$ over a set of frequencies (subcarriers) to undo the distortion and decode the data. These noisy, per-packet channel estimates are called channel state information (CSI).

\wifi sensing repurposes the need for channel estimation into a measurement tool. Since channel estimates are computed anyway for every packet to make communication possible, sensing methods simply leverage them as an observation of the physical environment. Because CSI captures how the environment shapes the signal, motion induces structured temporal changes in CSI that sensing methods learn to associate with properties of the underlying scene. \Cref{fig:csi_compare} illustrates this effect, where CSI magnitudes over time show a different pattern when a person walks through the room.
While CSI has traditionally been kept inside the receiver, recent research tools expose it on selected chipsets \cite{intel5300-paper,atheros-driver,esp-driver,nexmon-csi:2019, nexmon:2017}, and the upcoming IEEE 802.11bf standard \cite{ieee80211bf} further standardizes access to channel measurements for sensing.

\section{\wiflow Dataset} \label{subsec:dataset}
\begin{table}[t]
    \centering
    \caption{\textbf{Dataset comparison} between existing datasets and our proposed dataset based on key requirements for optical flow estimation. (\protect\checkmark) denotes datasets that are accessible and provide RGB frames, from which an optical flow \pseudo can be derived using our method (\cf supplement).}
    
    \setlength{\tabcolsep}{0.1cm}%
    \small%
    \begin{tabularx}{\colwidth}{X c c c | c}
        \toprule
        Dataset & Optical Flow & Device-free & Multi-RxTx & Accessible \\
        \midrule
        MMFi~\cite{mmfi2023} & $(\checkmark)$ & $\checkmark$ & $\crossmark$ & $\checkmark$\\
        XRF55~\cite{XRF55} & $\crossmark$ & $\checkmark$ & $\checkmark$ & $\crossmark$\\
        SignFi~\cite{signfi} & $\crossmark$ & $\checkmark$ & $\crossmark$ & $\checkmark$\\
        DICHASUS / ESPARGOS~\cite{dichasus2021} & $(\checkmark)$ & $\crossmark$ & $\checkmark$ & $\checkmark$\\
        Person-in-WiFi~\cite{person-in-wifi:2019} & $\crossmark$ & $\checkmark$ & $\checkmark$ & $\crossmark$\\
        Person-in-WiFi3D~\cite{person-in-wifi-3d:2024} & $\crossmark$ & $\checkmark$ & $\checkmark$ & $\crossmark$\\
        WiMans~\cite{wimans2024} & $(\checkmark)$ & $\checkmark$ & $\crossmark$ & $\checkmark$\\
        \midrule
        Ours & $\checkmark$ & $\checkmark$ & $\checkmark$ & $\checkmark$\\
         \bottomrule
    \end{tabularx}

    \label{tab:datasetcomparison}
\end{table}

We aim to learn a mapping from captured \wifi channel estimates, that is, channel state information (CSI), to motion in the scene, represented as optical flow. Training a deep neural network for this task requires paired data, with CSI measurements aligned to optical-flow labels.

Our target scenario is a fixed indoor environment where transmitters and receivers remain static and subjects carry no \wifi devices, that is, device-free sensing. Reliable supervision further requires tight synchronization between CSI and a modality that can provide flow labels, such as cameras. We rely on multiple receive antennas because a single antenna provides an incomplete and ambiguous view of motion. Similar channel changes can be caused by motion in different parts of the room, and some motion directions may cause little to no change at all. Multiple antennas offer complementary vantage points that help resolve where motion occurred, consistent with prior work~\cite{widir:2016,witraj:2023,WiDar3}.

Existing datasets do not meet these requirements. Optical-flow datasets offer RGB frames with flow annotations~\cite{Geiger:2012:AWR,Butler:2012:NOS,springdata2023,Dosovitskiy:2015:FN, flyingthings:2016,Baker:2007:DEM} but contain no CSI. Conversely, CSI datasets typically lack synchronized video and therefore cannot provide optical-flow supervision. Even when video is available, existing CSI datasets summarized in \cref{tab:datasetcomparison} do not capture the CSI we need for thorough evaluation, \ie, multi-device, multi-antenna measurements at high sample rate and bandwidth, in a device-free indoor setup. We therefore collect a novel dataset tailored to optical flow estimation from CSI.

\subsubsection{Capture Setup.}

\begin{figure}[t]
    \centering
    \begin{minipage}[t]{0.48\textwidth}
        \centering
        \includegraphics[width=\linewidth]{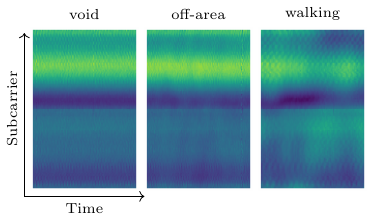}
        \caption{\textbf{Amplitude heatmaps} for multiple subcarriers over $100$ time-consecutive CSI for three different scenarios: No motion (void), motion only outside the captured area (off-area), and walking.}
        \label{fig:csi_compare}
    \end{minipage}
    \hfill
    \begin{minipage}[t]{0.48\textwidth}
        \includegraphics[width=\linewidth]{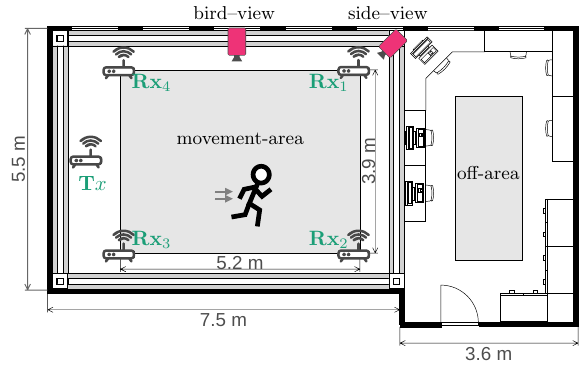}
        \caption{\textbf{Floorplan} of our capture setup showing the locations of routers, cameras, movement-, and \noise. All captured actions, except motion in off-area, are performed within the movement-area.}
        \label{fig:floorplan}
    \end{minipage}
\end{figure}

Following the practice of other \wifi sensing works ~\cite{mmfi2023,person-in-wifi:2019,person-in-wifi-3d:2024,hpe-li-pose:2024}, we assume a device-free setup, where locations of \wifi devices (transmitter and receivers), cameras, and walls remain fixed during capturing. Our experimental setup consists of one transmitter (Tx), four receivers (Rx$_1$ to Rx$_4$) placed in the corners of the movement area, and two cameras, as shown in \cref{fig:floorplan}.

To generate pseudo ground truth of optical flow, one camera captures the movement area from the side (\sideview), and the other from above (\birdview). We are only able to predict the motion of subjects visible in the cameras in the movement-area. Since movement outside the visible area (called \noise) also affects the wireless channel, we explicitly handle this case during data collection as well. In total, we collect data of seven predefined actions with one or two people moving, sequences without any motion (\void), and with motion only in the \noise. Each capture is \SI{30}{\second} long and repeated multiple times for each of the $10$ subjects. A detailed description of the actions, as well as exact statistics of the captured data, can be found in the supplemental \cref{suppl:action-details}. %

To capture high-frequency, high-bandwidth CSI, we use a single-antenna transmitter (USRP N2954-R) that broadcasts \wifi packets at \SI{1}{\kilo\hertz}. We operate on channel $157$ with \SI{80}{\mega\hertz} bandwidth, which is typically unused by nearby \wifi devices, reducing interference while maximizing the number of subcarriers and thus the richness of the CSI. We deploy four receivers (Asus RT-AC86U), each with four antennas, placed in the corners of the movement-area, and extract CSI using NexmonCSI~\cite{nexmon-csi:2019}. In total, we collect $328$ sequences, totaling \SI{164}{\minute}, and observe an average packet loss of only $0.5\%$ across devices and captures.

\subsubsection{Synchronizing Training Data.}

A key challenge in building training pairs is aligning the video frames with the much higher-rate CSI stream. Our two cameras capture video at $30$ Hz (side-view) and $50$ Hz (bird-view), while CSI is recorded at $1000$ Hz. We subsample the videos by keeping every fifth frame, which yields effective frame rates of $6$ Hz and $10$ Hz. Subsampling is needed to have a coarse enough temporal resolution for computing reasonable optical flow.
For each retained frame at time $t$, we then associate a fixed number of CSI measurements based on timestamp proximity, selecting the $K$ closest CSI samples \textit{prior} to $t$.

\begin{figure}[t]
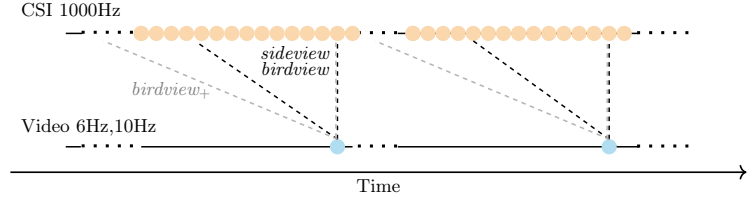

    \centering
    \includestandalone[width=.8\linewidth]{tikz/csi_video_matching}
    \caption{\textbf{Matching process} between CSI and video frames. For \sideview and \birdview we match the latest $10$ CSI with each image frame; for \birdviewplus we use the latest $100$.}
    \label{fig:csi_matching}
\end{figure}

We provide three aligned versions of the data. \textit{sideview} uses the side-view camera frames with $K=10$, following the pairing used in MM-Fi~\cite{mmfi2023}. \birdview uses the ceiling camera frames with the same $K=10$. To study the effect of larger CSI context per frame, we also introduce \birdviewplus, using the same frames as \birdview but with an increased associatied number of $K=100$ CSI. \Cref{fig:csi_matching} visualizes the resulting alignment.

\subsubsection{Pseudo Ground Truth.}
To acquire optical flow, we process the subsampled frames from both cameras using an ensemble of SOTA optical flow methods to produce pseudo ground truths (denoted as \pseudo) at a resolution of $168\times128$ pixels. 
Details on the generation of \pseudo can be found in the supplemental \cref{suppl:pseudo-detail}.

\begin{figure}[b]
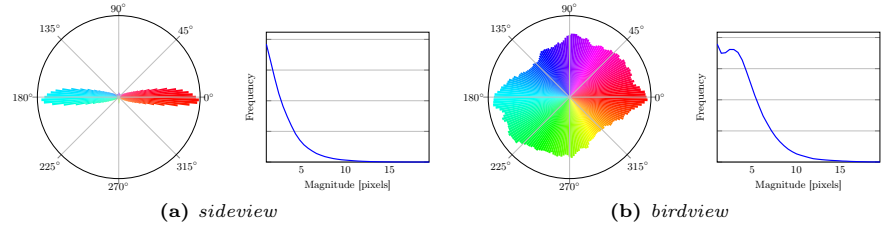

    \centering
    \begin{subfigure}[t]{0.48\linewidth}
        \centering
        \includestandalone[width=.95\linewidth]{tikz/angles07}
        \caption{\sideview}
    \end{subfigure}
    \begin{subfigure}[t]{0.48\linewidth}
        \centering
        \includestandalone[width=.95\linewidth]{tikz/angles08}
        \caption{\birdview}
    \end{subfigure}
    \caption{\textbf{Distributions} of the motion angles and magnitudes in our two dataset views of all moving pixels. The distribution of the motion angles reveals that most motions in the \sideview perspective are along the horizontal axis, while the motion directions in the \birdview perspective are more uniformly distributed. Similarly, there are more samples with faster motions in the \birdview perspective.}
    \label{fig:seemo-distr-polar}
\end{figure}

As shown in \cref{fig:seemo-distr-polar}, the distributions of motion angles differ drastically between the two camera views. Most motions in the \sideview are biased towards horizontal motions compared to the more uniformly distributed motions in \birdview.
Further, in both of our perspectives, slower motions are more common than faster ones, with the frequency gradually decaying. %

\subsubsection{Evaluation Settings.}

We introduce two different partitioning strategies of our dataset for model training and evaluation. In \textit{time split}, $3$ of the $4$ ($6$ of $8$, respectively) captured sequences per action are allocated for training, with the remaining reserved and split between validation and test. This allows us to assess temporal generalization while having the same subjects during training and testing. To evaluate cross-subject generalization, we additionally propose a \textit{subject split}, in which all actions and repetitions performed by seven persons are used for training, while the data of one other person is used for validation, and all actions of two other persons are used as a test set. See supplemental \cref{tab:split-detail} for further details.

\section{\wiflow}
We aim to reconstruct the motion field a camera would see, using only CSI from \wifi receivers. In general, CSI consists of complex-valued measurements $H_i\in\mathbb{C}^{T\times R\times D\times K\times N}$, with $T$ transmit antennas, $R$ antennas per receiver, $D$ receivers, $K$ subcarriers, and $N$ captured CSI snapshots. In our setup, $T=1$, $D=4$, and $R=4$; we stack all antenna/subcarrier measurements across the four receivers into a single antenna-like dimension $A=T\cdot D\cdot R$, and use a window of $N$ snapshots (either $10$ or $100$, see \cref{subsec:dataset}), yielding $H_i'\in\mathbb{C}^{A\times K\times N}$. The target is an optical-flow field $F_i\in\mathbb{R}^{2\times H\times W}$, where $H$ and $W$ are the spatial height and width, representing the motion that would be visible between two consecutive camera frames $t_i$ and $t_{i+1}$.

Our goal is to train a neural network that maps CSI sequences to a 2D optical flow field. This mapping is non-trivial because the input is a structured time--frequency tensor, while the output is a dense spatial motion field. We achieve this by leveraging existing CSI preprocessing to obtain a suitable representation utilized by our proposed model architectures tailored to this task.

\subsubsection{Model Architectures.}
\begin{figure}[t]
    \centering
    \includegraphics[width=.8\linewidth]{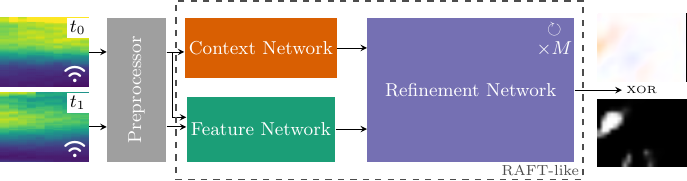}
    \caption{\textbf{\wiflow block} used in all of our architectures. First, the CSI data is preprocessed, and then we use a RAFT-like architecture to predict either optical flow or a motion mask.}
    \label{fig:block}
\end{figure}

\begin{figure}[t]
    \centering
    \includegraphics[width=.8\linewidth]{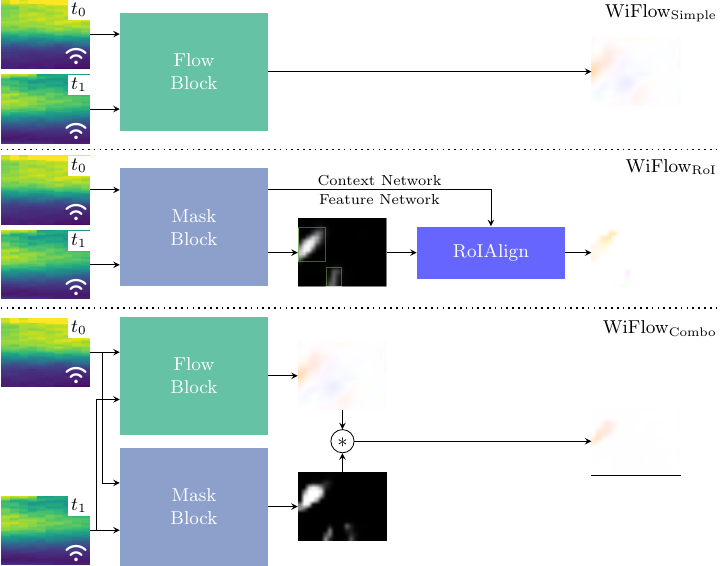}
    \caption{\textbf{Architectures.} In each of our proposed architectures, we use our \wiflow building block. \ours consists only of a flow block, while \ourm uses a mask block which additionally returns the features of its context and feature networks. The mask and additional features are then utilized by RoIAlign to calculate the optical flow prediction. In \ourl, we combine the outputs of a flow and mask block.}
    \label{fig:architectures}
\end{figure}

We propose three different models to predict optical flow from CSI. All of them utilize a basic building block inspired by RAFT~\cite{RAFT}, a state-of-the-art optical flow method for images. Our basic building block, as shown in \cref{fig:block}, combines a CSI preprocessor with a RAFT-like architecture to either directly predict optical flow or a motion mask.
Depending on the output, we refer to this basic building block as \textit{Flow Block} or \textit{Mask Block}, respectively.

As in RAFT, we use three different sub-networks: A \textit{Feature Network} that extracts per-pixel features from both input frames and calculates the similarity between these features, a \textit{Context Network} that extracts additional information from the input frames, and a \textit{Refinement Network} that combines the outputs of the other two. The \textit{Refinement Network} iteratively updates the prediction multiple times. %
In our setup, the context network and feature network are both based on a modified ResNet~\cite{RESNET} with an increased input dimensionality to match the dimensions of the preprocessor's output data, and the refinement network is the same convolutional GRU as in RAFT~\cite{RAFT, GRU}.

\cref{fig:architectures} visualizes our three model architectures of varying complexity. Our simplest baseline, \ours{}, consists of the flow block directly predicting the optical flow. 
Since devices in our setup are static and motion is sparse, large areas of the resulting flow are zero. To leverage this, we also design an architecture that separates the task into motion localization and motion estimation. \ourm{} is inspired by object detection and segmentation~\cite{maskrcnn:2017}. A mask block pretrained on CSI is used to predict moving areas. Then, only within the bounding boxes extracted from the motion mask contours, the optical flow is predicted using RoIAlign~\cite{maskrcnn:2017}. In our case, the RoIAlign receives bounding boxes and a combination of the features calculated in the context and feature networks of the mask block as inputs, and a convolutional decoder calculates the optical flow within each bounding box.
\ourl{} follows a similar motivation, but instead of sequentially performing motion localization and estimation, two branches perform these tasks in parallel. One branch computes the flow with a flow block, while the other branch predicts the motion mask with a mask block, respectively. The final result is obtained by point-wise multiplying the predicted flow with the motion mask. %

\subsubsection{Training Loss.}
Due to the static camera position, most pixels exhibit zero flow. To prevent training from collapsing to the trivial all-zero prediction, we modify the sequence loss of RAFT~\cite{RAFT} by up-weighting errors on non-zero flow pixels as
\begin{equation}
\mathcal{L} = \sum_{m=1}^{M} \gamma^{M-m} ( F_{gt} - F_m )^{4},
\label{eq:wiflow_loss}
\end{equation}
where $M$ is the number of iterations in the refinement network, $\gamma$ is the decay factor and $F_{m}$ is the optical flow prediction at the $m$-th refinement iteration.

\section{Experiments}
As no prior work has targeted the task of predicting optical flow directly from CSI, we conducted a detailed analysis of common CSI preprocessing methods and our proposed model architectures.

\subsection{Setup}
\subsubsection{Evaluation Metrics.}

\label{sec:metrics}
\begin{figure}[t]
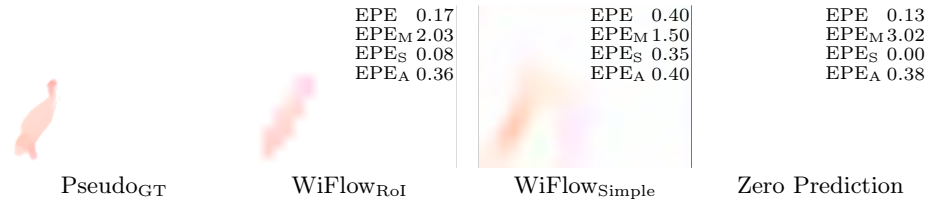

    \centering
    \includestandalone{tikz/metrics}
    \caption{\textbf{Metrics.} A comparison of our metrics on different predictions shows that for our setting, the EPE can be minimized when not predicting any motion (\emph{zero prediction}).
    We use EPE\textsubscript{A} as a more stable alternative. 
    Further, we evaluate moving and static areas separately (EPE\textsubscript{M} and EPE\textsubscript{S}).}
    \label{fig:metrics}
\end{figure}

Following common practice in optical flow~\cite{RAFT,SEARAFT,recover:2025}, we evaluate the endpoint-error (EPE), which measures the mean Euclidean norm of the differences between the predicted optical flow vectors and our pseudo ground truth (\pseudo). In our setup, due to the static camera, each frame of our dataset contains a large portion of static background without motion, biasing the overall EPE, as visualized in \cref{fig:metrics}. We therefore separately analyze the EPE for the moving and static pixels. EPE\textsubscript{M} refers to the EPE of the moving pixels, \ie pixels where the Euclidean norm of the \pseudo is larger than $0.5$, while EPE\textsubscript{S} refers to the EPE of the remaining static pixels. We report all three metrics in our experiments. We also introduce EPE\textsubscript{A}, where all areas are evaluated but amplified by $4$, similar to our loss. This amplifies the penalization of absent but expected motion. %

\subsubsection{Implementation Details.}
We train all our methods using a cosine annealing learning rate scheduler~\cite{SGDR:2017} with a base learning rate of $1e^{-3}$ on a single NVIDIA RTX 6000 Ada GPU. We set the decay factor of our loss function to $0.8$, and train all models using the AdamW optimizer~\cite{AdamW:2019} for $60k$ training steps with $4$ refinement iterations.

\subsection{Analysis}
\label{sec:ablations}

\begin{table}[t]
\begin{minipage}[t]{.45\linewidth}
    \centering
    \caption{\textbf{Preprocessing.} 
    Results of \ours trained on the time split of \birdviewplus{}. \textit{Zero} represents the baseline performance achieved by predicting constant-zero flows.}
    \begin{tabularx}{\linewidth}{X  S[table-format=1.2]  S[table-format=1.2]  S[table-format=1.2]  S[table-format=1.2]}
        \toprule
         & {EPE}          & {EPE\textsubscript{M}}      & {EPE\textsubscript{S}}     & {EPE\textsubscript{A}} \\
\midrule
\textit{Zero} &  0.17 &  3.70 &  0.00 &  0.84 \\
\midrule
SavGol &  0.59 &  3.64 &  0.44 &  1.14 \\

Fourier &  0.41 &  3.71 &  0.25 &  0.95 \\

Raw &  \bfseries 0.40 &  3.71 & \bfseries 0.24 &  0.94 \\

Quotient & 0.41        & \bfseries 2.89  & 0.29         & \bfseries 0.78\\

PCA &  0.47 &  2.95 &  0.35 &  0.82 \\
\bottomrule
\end{tabularx}
\label{tab:compare-preprocessors}
\end{minipage}
\hspace{0.6cm}
\begin{minipage}[t]{.45\textwidth}
    \setlength{\tabcolsep}{2.5pt}
    \centering
        \caption{\textbf{Computational requirements} for \birdviewplus during inference of our three proposed architectures averaged over $1000$ measurement runs on a single NVIDIA RTX 6000 Ada GPU.}
    \begin{tabularx}{\linewidth}{XS[table-format=2.0]S[table-format=2.0]S[table-format=3.0]}
            \toprule
             & {Time} & {FLOPs} & {Memory} \\[-1pt]
             & {in ms} & {$\times 10^{9}$} & {in MB} \\
            \midrule
            \ours{}   &  \bfseries 23 & \bfseries 22& \bfseries 380\\
            \ourm{}   &  26& \bfseries 22 & \bfseries 380\\
            \ourl{}  &  47 &43 & 763\\
    \bottomrule
    \end{tabularx}

    \label{tab:compare-inference}
\end{minipage}
\end{table}

\subsubsection{Preprocessing.}

We consider five CSI preprocessing strategies. \textit{Raw} uses amplitude and phase for every complex CSI value; amplitude is normalized to $[0,1]$. \textit{Quotient}~\cite{witraj:2023} normalizes across antennas: for each device, subcarrier and time step, we divide the complex measurement of each antenna by the measurement of a reference antenna. This largely cancels phase offsets shared within a device. \textit{Fourier} applies an inverse Fourier transform along the time dimension independently for each antenna and subcarrier, and uses the resulting real and imaginary parts as features. \textit{SavGol} smooths the real and imaginary parts of the CSI independently over time using a Savitzky--Golay filter (window length $51$). \textit{PCA} performs PCA-based denoising over the flattened per-timestep CSI vector ($A\times K \times 2$ for real/imag), keeping $150$ principal components; the PCA projection is computed once on the training set. The PCA features are projected back to the original space to preserve the input tensor shape.

\cref{tab:compare-preprocessors} shows the effect of the different preprocessing of CSI when training our \ours{}. 
Although \emph{Raw} achieves the lowest EPE, it does not achieve this for the actual moving pixels. For our setup, the EPE can be minimized by solely predicting zero optical flow values. 
\textit{Quotient} achieves the lowest EPE for the moving pixels (EPE\textsubscript{M}), while having also competitive low EPE for the background area (EPE\textsubscript{S}). This is further confirmed with its lowest EPE\textsubscript{A}. %
The quantitative gap between \emph{Raw} and \emph{Quotient} indicates that preprocessing selection is necessary and beneficial before further processing CSI with a neural network. 
Based on these findings, we train all subsequent models using the \textit{Quotient} preprocessor. %

\subsubsection{Impact of Device Count.}
\begin{figure}[t]
    \centering
    \begin{tikzpicture}

    \pgfplotstableread[col sep=comma]{
devices,epe,epem,epes,epea
4,0.43782032118112446,2.9432883589887973,0.3148829319844861,0.8061014603494965
3,0.4412350993706259,2.906410302971559,0.3203314152731428,0.7962159782864007
2,0.5495797102905212,3.170796801632372,0.42113310021023087,0.9611167892370027
1,0.5495797102905212,3.170796801632372,0.42113310021023087,0.9611167892370027
0.5,0.5466583693146592,3.346236214985734,0.40902492936370943,1.0362184069695675
    }\epe

    \begin{groupplot}[
        group style={
            group size=1 by 2,
            vertical sep=5pt,
            x descriptions at=edge bottom,
        },
        width=120mm,
        xmin=0.25, xmax=4.25,
        xtick={0.5,1,2,3,4},
        xticklabels={
            (1;\,2), (1;\,4), (2;\,4), (3;\,4), (4;\,4)
        },
        ymajorgrids,
        cycle list/Dark2,
        legend style={
            font=\scriptsize, 
            cells={anchor=west}, 
            at={(0.5,1.8)},
            legend columns=3,
            anchor=center
        },
    ]
        \nextgroupplot[
            height=2.5cm, 
            ymin=2.8, ymax=3.45,
            ytick={3.0, 3.2, 3.4},
            after end axis/.code={
                \draw [thick, white] (rel axis cs:0,0) -- (rel axis cs:1,0);
                \foreach \x in {0,1} {
                    \draw [thick] (rel axis cs:\x,0) +(-3pt,-2pt) -- +(3pt,2pt);
                }
            },
            axis x line*=top, 
            xtick=\empty,
        ]
        \addplot+[line width=0pt, mark=*] table [x=devices, y=epe] {\epe};
        \addplot+[line width=0pt, mark=*] table [x=devices, y=epem] {\epe};
        \addplot+[line width=0pt, mark=*] table [x=devices, y=epes] {\epe};

        \nextgroupplot[
            height=3cm,
            ymin=0.25, ymax=0.65,
            ytick={0.3,0.4,0.5, 0.6},
            xlabel={Devices;\,Antennas per device},
            legend entries={$\text{EPE}$, $\text{EPE}_\text{M}$, $\text{EPE}_\text{S}$, $\text{EPE}_\text{A}$},
            after end axis/.code={
                \foreach \x in {0,1} {
                    \draw [thick] (rel axis cs:\x,1) +(-3pt,-2pt) -- +(3pt,2pt);
                }
            },
            axis x line*=bottom
        ]
        \addplot+[line width=0pt, mark=*] table [x=devices, y=epe] {\epe};
        \addplot+[line width=0pt, mark=*] table [x=devices, y=epem] {\epe};
        \addplot+[line width=0pt, mark=*] table [x=devices, y=epes] {\epe};
    \end{groupplot}

    \node[rotate=90] at ($(group c1r1.west)!0.5!(group c1r2.west) + (-1cm,0)$) {Metric value};
\end{tikzpicture}
    \caption{\textbf{Impact of device count.} Accuracy on \birdviewplus{} when increasing the number of receiver devices (4 antennas per device). For the single-device setting, we also report a variant using only two antennas from the first receiver.}
    \label{fig:device-demand}
\end{figure}
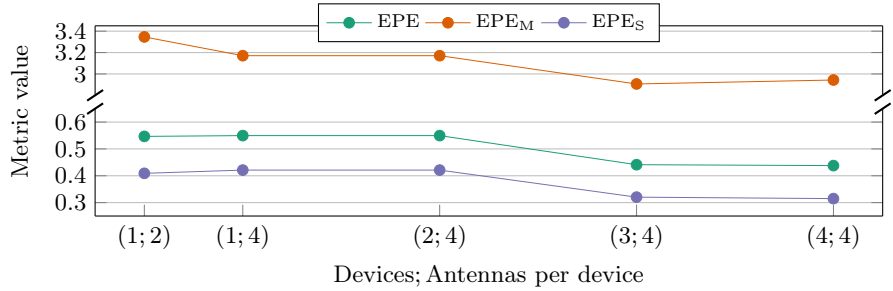

To evaluate the impact of the number of devices, we train multiple \ours{} models with $1$--$4$ receivers. For the single device setting, we additionally evaluate a $2$ antenna for a single receiver. As shown in \cref{fig:device-demand}, we observe improvements across all metrics as we increase the number of devices up to $3$, with only minor improvements beyond that. We therefore conclude that incorporating measurements from multiple spatially separated receiver locations is beneficial for performing spatial inference tasks such as optical flow estimation.
For further experiments, we use CSI from all $16$ antennas of the four receivers.

\subsubsection{Comparing Architectures.}
Our three architecture vary in computational requirements.  \cref{tab:compare-inference} shows that that \ours and \ourm have very similar requirements. \ourl requires roughly twice as many compute resources as the other methods due to having two of the \wiflow blocks. However, all of our methods require less than $1GB$ of VRAM, which allows running our models even on cheap consumer GPUs and inference is in the order of ms.

\begin{figure}[hbt]
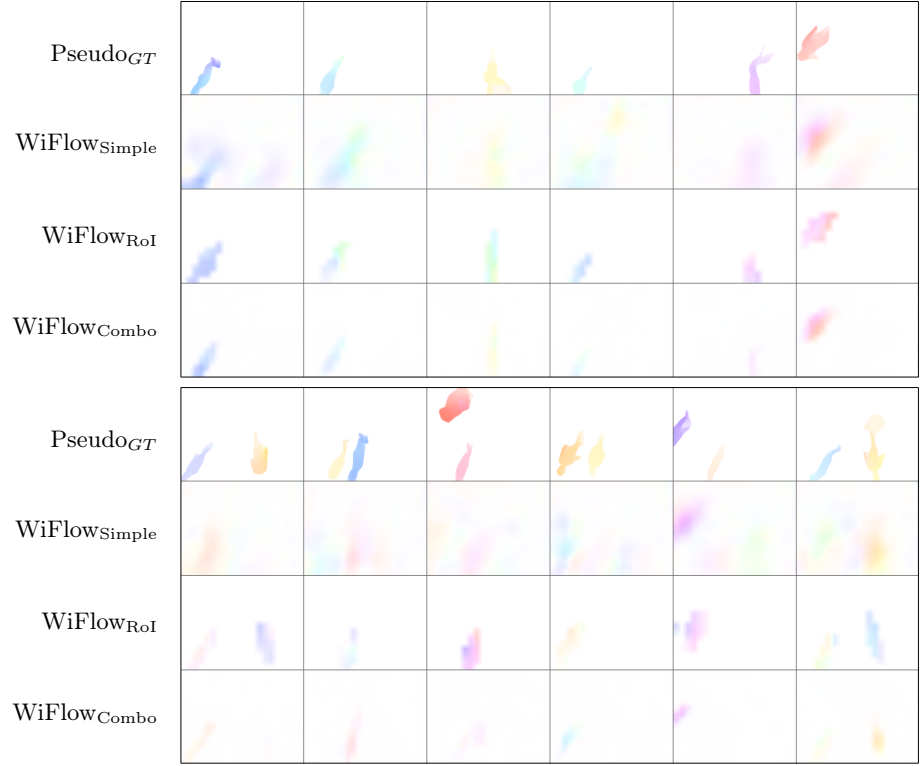

\centering
\includestandalone{tikz/qualitative}

\vspace{0.1cm}

\includestandalone{tikz/qualitative_together}
\caption{\textbf{Qualitative results} of each architecture trained on the time split of \birdviewplus. The upper half shows a single moving person in each scene, while the lower half shows samples with multiple moving persons.
\ours tends to predict more motions in the static background than our other models, and the overall prediction quality of our methods is lower when multiple persons are moving.}
\label{fig:qualitative-results}
\end{figure}

\cref{tab:compare-epe} shows the resulting accuracies of our proposed models on each perspective and evaluation split of our dataset. The results are consistent across all possible settings of our dataset and they show that \ourl performs best across most of our metrics. The only exception is EPE\textsubscript{M} where \ours outperforms all other models. However, the qualitative results, shown in \cref{fig:qualitative-results}, indicate that \ours suffers from a lot of noise in the non-moving background pixels, whereas our other models are better at localizing motion.
We provide further in-depth analysis of the accuracy of our networks on the different actions present in our dataset in supplemental \cref{suppl:per-action}.

\begin{table}[t!]
\centering
\caption{\textbf{Evaluation} of EPE ($\downarrow$) metrics of \wiflow on each evaluation setting of our proposed dataset.
}

    \setlength{\tabcolsep}{0.15cm}%
    \footnotesize%
    \begin{tabularx}{\linewidth}{@{}XXS[table-format=1.2]S[table-format=1.2]S[table-format=1.2]cS[table-format=1.2]S[table-format=1.2]S[table-format=1.2]@{}}
        \toprule
                                 &              & \multicolumn{3}{c}{Subject Split} & & \multicolumn{3}{c}{Time Split} \\
                                 \cmidrule(lr){3-5} \cmidrule(lr){7-9}
        \raisebox{2.5pt}[0pt][0pt]{\multirow{-2}{*}{Perspective}}              & \raisebox{2.5pt}[0pt][0pt]{\multirow{-2}{*}{Architecture}} & {EPE\hphantom{\textsubscript{M}}}          & {EPE\textsubscript{M}}     & {EPE\textsubscript{S}}   &   & {EPE\hphantom{\textsubscript{M}}}          & {EPE\textsubscript{M}}      & {EPE\textsubscript{S}}      \\
        \midrule
        \multirow{3}{*}{\sideview}     & \ours  & 0.51        & \bfseries 2.22 & 0.40      &    & 0.53        & \bfseries 2.07 & 0.43         \\
                                       & \ourm  & 0.18        & 2.36          & 0.04     &    & 0.16        & 2.21          & 0.04         \\
                                       & \ourl  & \bfseries 0.16 & 2.36        & \bfseries 0.02 & & \bfseries 0.15 & 2.18        & \textbf{0.03} \\
        \midrule
        \multirow{3}{*}{\birdview}     & \ours  & 0.55        & \bfseries 3.53 & 0.40     &     & 0.56        & \bfseries 3.24 & 0.43         \\
                                       & \ourm  & 0.21        & 3.73          & 0.05   &      & 0.20         & 3.40           & 0.04         \\
                                       & \ourl  & \bfseries 0.19 & 3.71        & \bfseries 0.03 &  & \bfseries 0.18 & 3.38        & \textbf{0.03} \\
        \midrule
        \multirow{3}{*}{\birdviewplus}  & \ours  & 0.45        & \bfseries 3.24 & 0.32    &     & 0.41        & \bfseries 2.89 & 0.29         \\
                                       & \ourm  & 0.21        & 3.41          & 0.05    &     & 0.19        & 3.13          & 0.05         \\
                                       & \ourl  & \bfseries 0.18 & 3.50         & \bfseries 0.02 & & \bfseries 0.17 & 3.13        & \textbf{0.02} \\
        \bottomrule
    \end{tabularx}

\label{tab:compare-epe}
\end{table}

\subsubsection{Cross Subject Generalization.}
In the evaluation of our two different split protocols, shown in \cref{tab:compare-epe}, we can see that the overall accuracy is very similar between time and subject split. This shows that all our architectures can generalize to unseen persons and that they do not overfit to the training data.

\subsubsection{Impact of Output Resolution.}
\begin{table}[t]
\centering
\caption{\textbf{Resolution.} Evaluation of EPE ($\downarrow$) metrics on different resolutions on the subject split of \birdviewplus.}

    \setlength{\tabcolsep}{0.15cm}%
    \footnotesize%
    \begin{tabularx}{\colwidth}{@{}XS[table-format=1.2]S[table-format=1.2]S[table-format=1.2]cS[table-format=1.2]S[table-format=1.2]S[table-format=1.2]@{}}
        \toprule
                                  & \multicolumn{3}{c}{$128 \times 168$} & & \multicolumn{3}{c}{$256\times 336$} \\
                                 \cmidrule(lr){2-4} \cmidrule(lr){6-8}
        \raisebox{2.5pt}[0pt][0pt]{\multirow{-2}{*}{Architecture}} & {EPE\hphantom{\textsubscript{M}}}          & {EPE\textsubscript{M}}     & {EPE\textsubscript{S}}   &   & {EPE\hphantom{\textsubscript{M}}}          & {EPE\textsubscript{M}}      & {EPE\textsubscript{S}}      \\
        \midrule
        \ours  & 0.45        & \bfseries 3.24 & 0.32    &     &  0.87 & \bfseries 7.00 & 0.59 \\
        \ourm  & 0.21        & 3.41          & 0.05    &     & 0.42 & 7.40 & 0.10\\
        \ourl  & \bfseries 0.18 & 3.50         & \bfseries 0.02 & &\bfseries 0.37 & 7.41 & \bfseries 0.05\\
        \bottomrule
    \end{tabularx}

\label{tab:compare-resolution}
\end{table}

To explore the impact of optical flow image resolution on our models, we additionally train all models with a target resolution of $256 \times 336$ pixels, rather than the $128 \times 168$ pixels as used in our other experiments. In \cref{tab:compare-resolution}, we find that all of the measured EPEs roughly double when doubling the resolution, meaning that the overall error is comparable and dominated by the capabilities of the prediction from CSI-data.

\subsubsection{Limitations.}
\label{sec:limit}
We have shown that our framework is capable of learning to predict optical flow directly from CSI. However, similar to other methods in the field of \wifi sensing, \wiflow is limited to the environment trained for and does not generalize to other rooms or environments. Additional limitations arise from the fact that we do not have exact ground truth for the motion. 
Specifically, the \pseudo of our proposed dataset suffers from inconsistencies regarding the motion of shadows. While optical flow tracks the movement of the shadows, these motions cannot be tracked in the frequency domain of the CSI. %
Further, we have shown preliminary results that our framework can also generalize to multi-person movements in \cref{fig:qualitative-results}. However, it is not as precise as for single-subject actions, most likely due to the low percentage of multi-person data in the dataset. In addition, often multiple people move in different directions. Although \ourm supports multiple bounding boxes, it will struggle with occluded areas by design.

Another limitation we found is that the output resolution and level of detail of our models are relatively low compared to optical flow models that use camera inputs. While this limitation is expected, since CSI cannot be directly compared with camera frames, it may limit the practical applicability of our models in their current form. However, it also raises new research questions about how to increase the level of detail of these models, and we hope that this issue will be further explored in future work, building on the insights we gained.

\section{Conclusion}
In this work, we have demonstrated that we can estimate motion, in the form of optical flow, within a fixed scene from CSI sequences.
We proposed and compared three model architectures and found that our method \ourl, which leverages sharpening through a dedicated motion mask predictor, yielded the best accuracies. These results are consistent across different perspectives and even generalize to subjects not part of the training data.
Our novel sensing dataset is the first to incorporate optical flow as a modality and will be publicly released to encourage further research on motion estimation from CSI. 
With this, CSI-based optical flow estimation is a candidate technology for application scenarios where camera-based systems are unavailable or undesired.

{\small \section*{Acknowledgment}
We would like to thank G. Matthes for his technical support during data capture and M. Schulz \etal for publishing the Nexmon firmware~\cite{nexmon}, which enabled us to capture CSI from consumer devices in the first place.

SK has been funded by the Deutsche Forschungsgemeinschaft (DFG, German Research Foundation) under Germany´s Excellence Strategy --~EXC-3057.
FP has been funded by the European Union’s EU Framework Programme for Research and Innovation under the HORIZON-MSCA-DN-2022 Grant Agreement No.~101119652 (MSCA-DN-6th Sense).
MH was supported by the State of Hesse through LOEWE emergenCITY (Grant no. LOEWE/1/12/519/03/05.001(0016)/72).
SSM has been funded by the DFG --~project No.~529680848.
Calculations for this research were conducted on the Lichtenberg high-performance computer of the TU Darmstadt.
}

\bibliographystyle{splncs04}
\bibliography{bib/custom_length,bib/papers,bib/external,bib/csi,bib/cv}

\begin{thebibliography}{10}
\providecommand{\url}[1]{\texttt{#1}}
\providecommand{\urlprefix}{URL }
\providecommand{\doi}[1]{https://doi.org/#1}

\bibitem{ieee80211bf}
{IEEE} standard for information technology---{T}elecommunications and
  information exchange between systems local and metropolitan area
  networks---{S}pecific requirements---{P}art 11: {W}ireless lan medium access
  control ({MAC}) and physical layer ({PHY}) specifications---{A}mendment 4:
  {E}nhancements for wireless {LAN} sensing. {IEEE Std 802.11bf-2025 (Amendment
  to IEEE 802.11-2024, as amended by IEEE 802.11bh-2024, IEEE 802.11be-2024,
  and IEEE 802.11bk-2025)} pp. 1--228 (2025).
  \doi{10.1109/IEEESTD.2025.11184214}

\bibitem{wikey:2015}
Ali, K., Liu, A.X., Wang, W., Shahzad, M.: Keystroke recognition using {WiFi}
  signals. In: ACM SIGMOBILE International Conference on Mobile Computing and
  Networking. p. 90–102 (2015). \doi{10.1145/2789168.2790109}

\bibitem{csi2depth:2025}
{\'A}lvarez~Casado, C., Lage~Ca{\~{n}}ellas, M., Mustaniemi, J., Pedone, M.,
  Silv{\'e}n, O., Bordallo~L{\'o}pez, M.: {CSI2Depth}: {S}patio-temporal depth
  images from {Wi-Fi} {CSI} data via transformer networks and conditional
  generative adversarial networks. In: Image Analysis. pp. 368--382 (2025).
  \doi{10.1007/978-3-031-95911-0\_26}

\bibitem{esp-driver}
Atif, M., Muralidharan, S., Ko, H., Yoo, B.: {Wi-ESP—A} tool for {CSI}-based
  device-free {Wi-Fi} sensing ({DFWS}). Journal of Computational Design and
  Engineering  \textbf{7}(5),  644--656 (2020). \doi{10.1093/jcde/qwaa048}

\bibitem{Baker:2007:DEM}
Baker, S., Scharstein, D., Lewis, J.P., Roth, S., Black, M.J., Szeliski, R.: A
  database and evaluation methodology for optical flow. In: IEEE/CVF
  International Conference on Computer Vision. pp.~1--8 (2007).
  \doi{10.1109/ICCV.2007.4408903}

\bibitem{Black:1996:REM}
Black, M.J., Anandan, P.: The robust estimation of multiple motions:
  {P}arametric and piecewise-smooth flow fields. Computer Vision and Image
  Understanding  \textbf{63}(1),  75--104 (1996). \doi{10.1006/cviu.1996.0006}

\bibitem{Butler:2012:NOS}
Butler, D.J., Wulff, J., Stanley, G.B., Black, M.J.: A naturalistic open source
  movie for optical flow evaluation. In: European Conference on Computer
  Vision. vol.~7577, pp. 611--625 (2012). \doi{10.1007/978-3-642-33783-3_44}

\bibitem{wivelo:2024}
Cao, Z., Li, C., Liu, L., Zhang, M.: {WiVelo}: {F}ine-grained {Wi-Fi} walking
  velocity estimation. ACM Transactions on Sensor Networks  \textbf{20}(4)
  (2024). \doi{10.1145/3664196}

\bibitem{rfinpainter:2022}
Chen, C., Nishio, T., Bennis, M., Park, J.: {RF-Inpainter}: {M}ultimodal image
  inpainting based on vision and radio signals. IEEE Access  \textbf{10},
  110689--110700 (2022). \doi{10.1109/ACCESS.2022.3214972}

\bibitem{csiinpainter:2025}
Chen, C., Ohta, S., Nishio, T., Bennis, M., Park, J., Wahib, M.: Enabling
  visual scene recovery from {Wi-Fi} {CSI} for occlusion-free surveillance.
  IEEE Internet of Things Journal  \textbf{12}(11),  15040--15056 (2025).
  \doi{10.1109/JIOT.2025.3529499}

\bibitem{6GObjectTrackRobo:2023}
Chen, T., Gu, D.: {6D} object pose tracking with optical flow network for
  robotic manipulation. IFAC-PapersOnLine  \textbf{56}(2),  8048--8053 (2023).
  \doi{https://doi.org/10.1016/j.ifacol.2023.10.930}

\bibitem{GRU}
Cho, K., van Merrienboer, B., Bahdanau, D., Bengio, Y.: On the properties of
  neural machine translation: {E}ncoder-decoder approaches. In: {EMNLP Workshop
  on Syntax, Semantics and Structure in Statistical Translation}. pp. 103--111
  (2014). \doi{10.3115/V1/W14-4012}

\bibitem{Choudhury:2022:GWM}
Choudhury, S., Karazija, L., Laina, I., Vedaldi, A., Rupprecht, C.: Guess what
  moves: {U}nsupervised video and image segmentation by anticipating motion.
  In: British Machine Vision Conference (2022)

\bibitem{hpe-li-pose:2024}
D.~Gian, T., Dac~Lai, T., Van~Luong, T., Wong, K.S., Nguyen, V.D.: {HPE-Li}:
  {WiFi}-enabled lightweight dual selective kernel convolution for human pose
  estimation. In: European Conference on Computer Vision. vol. 15089, pp.
  93--111 (2024). \doi{10.1007/978-3-031-72751-1_6}

\bibitem{Dong:2023:VFI}
Dong, J., Ota, K., Dong, M.: Video frame interpolation: {A} comprehensive
  survey. {ACM Transactions on Multimedia Computing, Communications, and
  Applications}  \textbf{19}(78),  2s:1--2s:31 (2023). \doi{10.1145/3556544}

\bibitem{Memflow2024}
Dong, Q., Fu, Y.: {MemFlow}: Optical flow estimation and prediction with
  memory. In: IEEE/CVF Conference on Computer Vision and Pattern Recognition.
  pp. 19068--19078 (2024). \doi{10.1109/CVPR52733.2024.01804}

\bibitem{Dosovitskiy:2015:FN}
Dosovitskiy, A., Fischer, P., Ilg, E., H{\"a}usser, P., Haz{\i}rba\c{s}, C.,
  Golkov, V., {v. d. Smagt}, P., Cremers, D., Brox, T.: {FlowNet}: {L}earning
  optical flow with convolutional networks. In: IEEE/CVF International
  Conference on Computer Vision. pp. 2758--2766 (2015).
  \doi{10.1109/ICCV.2015.316}

\bibitem{dichasus2021}
Euchner, F., Gauger, M., D\"orner, S., ten Brink, S.: A distributed massive
  {MIMO} channel sounder for ``big {CSI} data''-driven machine learning. In:
  International ITG Workshop on Smart Antennas (2021)

\bibitem{Gao:2020:FGV}
Gao, C., Saraf, A., Huang, J., Kopf, J.: Flow-edge guided video completion. In:
  European Conference on Computer Vision. vol. 12357, pp. 713--729 (2020).
  \doi{10.1007/978-3-030-58610-2\_42}

\bibitem{Geiger:2012:AWR}
Geiger, A., Lenz, P., Urtasun, R.: Are we ready for autonomous driving? {T}he
  {KITTI} vision benchmark suite. In: IEEE/CVF Conference on Computer Vision
  and Pattern Recognition. pp. 3354--3361 (2012).
  \doi{10.1109/CVPR.2012.6248074}

\bibitem{nexmon-csi:2019}
Gringoli, F., Schulz, M., Link, J., Hollick, M.: Free your {CSI}: {A} channel
  state information extraction platform for modern {Wi-Fi} chipsets. In:
  International Workshop on Wireless Network Testbeds, Experimental Evaluation
  \& CHaracterization. p. 21–28 (2019). \doi{10.1145/3349623.3355477}

\bibitem{csipose:2025}
Gu, Y., Chen, J., Chen, C., He, K., Jia, J., Feng, Y., Du, R., Wu, C.:
  {CSIPose}: {U}nveiling human poses using commodity {WiFi} devices through the
  wall. IEEE Transactions on Mobile Computing  \textbf{24}(10),  10914--10926
  (2025). \doi{10.1109/TMC.2025.3571469}

\bibitem{csi-gesture}
{H. {Xiong} and F. {Gong} and L. {Qu} and C. {Du} and K. {Harfoush}}:
  {CSI}-based device-free gesture detection. In: {International Conference on
  High-capacity Optical Networks and Enabling/Emerging Technologies}.
  pp.~{1--5} ({2015}). \doi{10.1109/HONET.2015.7395430}

\bibitem{Hahn:2025:SCU}
Hahn, O., Reich, C., Araslanov, N., Cremers, D., Rupprecht, C., Roth, S.:
  Scene-centric unsupervised panoptic segmentation. In: IEEE/CVF Conference on
  Computer Vision and Pattern Recognition. pp. 24485--24495 (2025).
  \doi{10.1109/CVPR52734.2025.02280}

\bibitem{intel5300-paper}
Halperin, D., Hu, W., Sheth, A., Wetherall, D.: Tool release: {G}athering
  802.11n traces with channel state information. ACM SIGCOMM Computer
  Communication Review  \textbf{41}(1), ~53 (2011).
  \doi{10.1145/1925861.1925870}

\bibitem{maskrcnn:2017}
He, K., Gkioxari, G., Dollár, P., Girshick, R.: {Mask R-CNN}. In: IEEE/CVF
  International Conference on Computer Vision. pp. 2980--2988 (2017).
  \doi{10.1109/ICCV.2017.322}

\bibitem{RESNET}
He, K., Zhang, X., Ren, S., Sun, J.: Deep residual learning for image
  recognition. In: IEEE/CVF Conference on Computer Vision and Pattern
  Recognition. pp. 770--778 (2016). \doi{10.1109/CVPR.2016.90}

\bibitem{Drone2024}
Ho, H.W., Zhou, Y., Feng, Y., de~Croon, G.C.H.E.: Optical flow-based control
  for micro air vehicles: an efficient data-driven incremental nonlinear
  dynamic inversion approach. Autonomous Robots  \textbf{48}(8) (2024).
  \doi{10.1007/s10514-024-10174-4}

\bibitem{Horn:1981:DOF}
Horn, B.K.P., Schunck, B.G.: Determining optical flow. Artificial Intelligence
  \textbf{17}(1--3),  185--203 (1981). \doi{10.1016/0004-3702(81)90024-2}

\bibitem{wipe:2025}
Huang, H., Wang, P., Zhao, L., Dai, Z., Liu, G., Gao, H.: {WiPE}:
  {P}rivacy-friendly {WiFi}-based human pose estimation on consumer platform.
  IEEE Transactions on Consumer Electronics  \textbf{71}(2),  5127--5135
  (2025). \doi{10.1109/TCE.2025.3547393}

\bibitem{wimans2024}
Huang, S., Li, K., You, D., Chen, Y., Lin, A., Liu, S., Li, X., McCann, J.A.:
  {WiMANS}: {A} benchmark dataset for {WiFi}-based multi-user activity sensing.
  In: European Conference on Computer Vision. pp. 72--91 (2024).
  \doi{10.1007/978-3-031-72946-1_5}

\bibitem{Hur:2019:IRR}
Hur, J., Roth, S.: Iterative residual refinement for joint optical flow and
  occlusion estimation. In: IEEE/CVF Conference on Computer Vision and Pattern
  Recognition. pp. 5754--5763 (2019). \doi{10.1109/CVPR.2019.00590}

\bibitem{Ilg:2017:FN2}
Ilg, E., Mayer, N., Saikia, T., Keuper, M., Dosovitskiy, A., Brox, T.: {FlowNet
  2.0}: {E}volution of optical flow estimation with deep networks. In: IEEE/CVF
  Conference on Computer Vision and Pattern Recognition. pp. 1647--1655 (2017).
  \doi{10.1109/CVPR.2017.179}

\bibitem{gma:2021}
Jiang, S., Campbell, D., Lu, Y., Li, H., Hartley, R.: Learning to estimate
  hidden motions with global motion aggregation. In: IEEE/CVF International
  Conference on Computer Vision. pp. 9772--9781 (2021).
  \doi{10.1109/ICCV48922.2021.00963}

\bibitem{wipose:2020}
Jiang, W., Xue, H., Miao, C., Wang, S., Lin, S., Tian, C., Murali, S., Hu, H.,
  Sun, Z., Su, L.: Towards {3D} human pose construction using {WiFi}. In: ACM
  SIGMOBILE International Conference on Mobile Computing and Networking. pp.
  1--14 (2020). \doi{10.1145/3372224.3380900}

\bibitem{jindirection:2020}
Jin, Y., Tian, Z., Li, Y., Li, Z., Zhang, Z.: A novel device-free tracking
  system using {WiFi}: {T}urning fading channel from foe to friend. In: IEEE
  International Conference on Communications. pp.~1--6 (2020).
  \doi{10.1109/ICC40277.2020.9148609}

\bibitem{csi2image:2021}
Kato, S., Fukushima, T., Murakami, T., Abeysekera, H., Iwasaki, Y., Fujihashi,
  T., Watanabe, T., Saruwatari, S.: {CSI2Image}: {I}mage reconstruction from
  channel state information using generative adversarial networks. IEEE Access
  \textbf{9},  47154--47168 (2021). \doi{10.1109/ACCESS.2021.3066158}

\bibitem{Ke:2021:OAV}
Ke, L., Tai, Y., Tang, C.: Occlusion-aware video object inpainting. In:
  IEEE/CVF International Conference on Computer Vision. pp. 14448--14458
  (2021). \doi{10.1109/ICCV48922.2021.01420}

\bibitem{wi2vi:2020}
Kefayati, M.H., Pourahmadi, V., Aghaeinia, H.: {Wi2Vi}: {G}enerating video
  frames from {WiFi} {CSI} samples. IEEE Sensors Journal  \textbf{20}(19),
  11463--11473 (2020). \doi{10.1109/JSEN.2020.2996078}

\bibitem{mwi2vi:2022}
Kefayati, M.H., Pourahmadi, V., Aghaeinia, H.: Multi-view {WiFi} imaging.
  Signal Processing  \textbf{197},  108552 (2022).
  \doi{10.1016/j.sigpro.2022.108552}

\bibitem{recover:2025}
Kiefhaber, S., Roth, S., Schaub-Meyer, S.: Removing cost volumes from optical
  flow estimators. In: IEEE/CVF International Conference on Computer Vision
  (2025)

\bibitem{pulsefi:2025}
Kocheta, P., Bhatia, N.S., Obraczka, K.: {Pulse-Fi}: {A} low-cost system for
  accurate heart rate monitoring using {Wi-Fi} channel state information. In:
  International Conference on Distributed Computing in Smart Systems and the
  Internet of Things. pp. 226--230 (2025).
  \doi{10.1109/DCOSS-IoT65416.2025.00037}

\bibitem{crowdcount:2018}
Li, J., Tu, P., Wang, H., Wang, K., Yu, L.: A novel device-free counting method
  based on channel status information. Sensors  \textbf{18}(11) (2018).
  \doi{10.3390/s18113981}

\bibitem{indotrack:2017}
Li, X., Zhang, D., Lv, Q., Xiong, J., Li, S., Zhang, Y., Mei, H.: {IndoTrack}:
  {D}evice-free indoor human tracking with commodity {Wi-Fi}. ACM Interactive,
  Mobile, Wearable, and Ubiquitous Technologies  \textbf{1}(3) (2017).
  \doi{10.1145/3130940}

\bibitem{SGDR:2017}
Loshchilov, I., Hutter, F.: {SGDR}: {S}tochastic gradient descent with warm
  restarts. In: International Conference on Learning Representations (2017)

\bibitem{AdamW:2019}
Loshchilov, I., Hutter, F.: Decoupled weight decay regularization. In:
  International Conference on Learning Representations (2019)

\bibitem{lucaskanade}
Lucas, B.D., Kanade, T.: An iterative image registration technique with an
  application to stereo vision. In: {International Joint Conference on
  Artificial Intelligence}. vol.~2, pp. 674--679 (1981),
  \url{https://hal.science/hal-03697340}

\bibitem{signfi}
Ma, Y., Zhou, G., Wang, S., Zhao, H., Jung, W.: {SignFi}: Sign language
  recognition using wifi. ACM Interactive, Mobile, Wearable, and Ubiquitous
  Technologies  \textbf{2}(1) (2018). \doi{10.1145/3191755}

\bibitem{flyingthings:2016}
Mayer, N., Ilg, E., H{\"a}usser, P., Fischer, P., Cremers, D., Dosovitskiy, A.,
  Brox, T.: A large dataset to train convolutional networks for disparity,
  optical flow, and scene flow estimation. In: IEEE/CVF Conference on Computer
  Vision and Pattern Recognition. pp. 4040--4048 (2016).
  \doi{10.1109/CVPR.2016.438}

\bibitem{springdata2023}
Mehl, L., Schmalfuss, J., Jahedi, A., Nalivayko, Y., Bruhn, A.: Spring: {A}
  high-resolution high-detail dataset and benchmark for scene flow, optical
  flow and stereo (2023). \doi{10.1109/CVPR52729.2023.00482}

\bibitem{DPFlow2025}
Morimitsu, H., Zhu, X., Cesar-Jr., R.M., Ji, X., Yin, X.C.: {DPFlow}: Adaptive
  optical flow estimation with a dual-pyramid framework. In: IEEE/CVF
  Conference on Computer Vision and Pattern Recognition. pp. 17810--17820
  (2025). \doi{10.1109/CVPR52734.2025.01659}

\bibitem{Niklaus:2018:CAS}
Niklaus, S., Liu, F.: Context-aware synthesis for video frame interpolation.
  In: IEEE/CVF Conference on Computer Vision and Pattern Recognition. pp.
  1701--1710 (2018). \doi{10.1109/CVPR.2018.00183}

\bibitem{SoftmaxSplatting:2020}
Niklaus, S., Liu, F.: Softmax splatting for video frame interpolation. In:
  IEEE/CVF Conference on Computer Vision and Pattern Recognition. pp.
  5436--5445 (2020). \doi{10.1109/CVPR42600.2020.00548}

\bibitem{PolarMendoza:2025}
Polar~Mendoza, E.G., Patiño, R., Cardinale, Y.: Social robot path planning
  based on a global perspective using optical flow. Journal of Intelligent {\&}
  Robotic Systems  \textbf{111}(4) (2025). \doi{10.1007/s10846-025-02323-3}

\bibitem{wisee:2013}
Pu, Q., Gupta, S., Gollakota, S., Patel, S.: Whole-home gesture recognition
  using wireless signals. In: ACM SIGMOBILE International Conference on Mobile
  Computing and Networking. p. 27–38 (2013). \doi{10.1145/2500423.2500436}

\bibitem{widar:2017}
Qian, K., Wu, C., Yang, Z., Liu, Y., Jamieson, K.: Widar: {D}ecimeter-level
  passive tracking via velocity monitoring with commodity {Wi-Fi}. In: ACM
  International Symposium on Mobile Ad Hoc Networking and Computing (2017).
  \doi{10.1145/3084041.3084067}

\bibitem{widar2:2018}
Qian, K., Wu, C., Zhang, Y., Zhang, G., Yang, Z., Liu, Y.: Widar2.0: {P}assive
  human tracking with a single {Wi-Fi} link. In: International Conference on
  Mobile Systems, Applications, and Services. pp. 350--361 (2018).
  \doi{10.1145/3210240.3210314}

\bibitem{widance:2017}
Qian, K., Wu, C., Zhou, Z., Zheng, Y., Yang, Z., Liu, Y.: Inferring motion
  direction using commodity {Wi-Fi} for interactive exergames. In: Conference
  on Human Factors in Computing Systems. p. 1961–1972 (2017).
  \doi{10.1145/3025453.3025678}

\bibitem{spynet:2017}
Ranjan, A., Black, M.J.: Optical flow estimation using a spatial pyramid
  network. In: IEEE/CVF Conference on Computer Vision and Pattern Recognition.
  pp. 2720--2729 (2017). \doi{10.1109/CVPR.2017.291}

\bibitem{winect:2022}
Ren, Y., Wang, Z., Tan, S., Chen, Y., Yang, J.: Winect: {3D} human pose
  tracking for free-form activity using commodity {WiFi}. ACM Interactive,
  Mobile, Wearable, and Ubiquitous Technologies  \textbf{5}(4) (2022).
  \doi{10.1145/3494973}

\bibitem{gopose:2022}
Ren, Y., Wang, Z., Wang, Y., Tan, S., Chen, Y., Yang, J.: {GoPose}: {3D} human
  pose estimation using {WiFi}. ACM Interactive, Mobile, Wearable, and
  Ubiquitous Technologies  \textbf{6}(2) (2022). \doi{10.1145/3534605}

\bibitem{Safadoust:2023:MOD}
Safadoust, S., G{\"{u}}ney, F.: Multi-object discovery by low-dimensional
  object motion. In: IEEE/CVF International Conference on Computer Vision. pp.
  734--744 (2023). \doi{10.1109/ICCV51070.2023.00074}

\bibitem{nexmon:2017}
Schulz, M., Wegemer, D., Hollick, M.: Nexmon: Build your own {Wi-Fi} testbeds
  with low-level mac and phy-access using firmware patches on off-the-shelf
  mobile devices. In: International Workshop on Wireless Network Testbeds,
  Experimental Evaluation \& CHaracterization. p. 59–66 (2017).
  \doi{10.1145/3131473.3131476}

\bibitem{nexmon}
Schulz, M., Wegemer, D., Hollick, M.: Nexmon: {T}he {C}-based firmware patching
  framework (2017), \url{https://nexmon.org}

\bibitem{flowformer++2023}
Shi, X., Huang, Z., Li, D., Zhang, M., Cheung, K.C., See, S., Qin, H., Dai, J.,
  Li, H.: {FlowFormer++}: Masked cost volume autoencoding for pretraining
  optical flow estimation. In: IEEE/CVF Conference on Computer Vision and
  Pattern Recognition. pp. 1599--1610 (2023).
  \doi{10.1109/CVPR52729.2023.00160}

\bibitem{Sim:2021:XVF}
Sim, H., Oh, J., Kim, M.: {XVFI:} {eX}treme video frame interpolation. In:
  IEEE/CVF International Conference on Computer Vision. pp. 14469--14478
  (2021). \doi{10.1109/ICCV48922.2021.01422}

\bibitem{Sun:2018:PWC}
Sun, D., Yang, X., Liu, M.Y., Kautz, J.: {PWC-Net}: {CNN}s for optical flow
  using pyramid, warping, and cost volume. In: IEEE/CVF Conference on Computer
  Vision and Pattern Recognition. pp. 8934--8943 (2018).
  \doi{10.1109/CVPR.2018.00931}

\bibitem{Sun:2020:MMS}
Sun, D., Yang, X., Liu, M., Kautz, J.: Models matter, so does training: {A}n
  empirical study of {CNNs} for optical flow estimation. IEEE Transactions on
  Pattern Analysis and Machine Intelligence pp. 1408--1423 (2020).
  \doi{10.1109/TPAMI.2019.2894353}

\bibitem{RAFT}
Teed, Z., Deng, J.: {RAFT}: {R}ecurrent all-pairs field transforms for optical
  flow. In: European Conference on Computer Vision. vol. 12347, p. 402–419
  (2020). \doi{10.1007/978-3-030-58536-5_24}

\bibitem{XRF55}
Wang, F., Lv, Y., Zhu, M., Ding, H., Han, J.: {XRF55}: {A} radio frequency
  dataset for human indoor action analysis. {ACM Interactive, Mobile, Wearable,
  and Ubiquitous Technologies}  \textbf{8},  21:1--21:34 (2024).
  \doi{10.1145/3643543}

\bibitem{person-in-wifi:2019}
Wang, F., Zhou, S., Panev, S., Han, J., Huang, D.: {Person-in-WiFi}:
  {F}ine-grained person perception using {WiFi}. In: IEEE/CVF International
  Conference on Computer Vision. pp. 5452--5461 (2019).
  \doi{10.1109/ICCV.2019.00555}

\bibitem{lifs:2016}
Wang, J., Jiang, H., Xiong, J., Jamieson, K., Chen, X., Fang, D., Xie, B.:
  {LiFS}: Low human-effort, device-free localization with fine-grained
  subcarrier information. In: ACM SIGMOBILE International Conference on Mobile
  Computing and Networking. pp. 243--256 (2016). \doi{10.1145/2973750.2973776}

\bibitem{carm:2015}
Wang, W., Liu, A.X., Shahzad, M., Ling, K., Lu, S.: Understanding and modeling
  of wifi signal based human activity recognition. In: ACM SIGMOBILE
  International Conference on Mobile Computing and Networking. p. 65–76
  (2015). \doi{10.1145/2789168.2790093}

\bibitem{deepfi:2017}
Wang, X., Gao, L., Mao, S., Pandey, S.: {CSI}-based fingerprinting for indoor
  localization: {A} deep learning approach. IEEE Transactions on Vehicular
  Technology  \textbf{66}(1),  763--776 (2017). \doi{10.1109/TVT.2016.2545523}

\bibitem{multimesh:2024}
Wang, Y., Ren, Y., Yang, J.: Multi-subject {3D} human mesh construction using
  commodity {WiFi}. ACM Interactive, Mobile, Wearable, and Ubiquitous
  Technologies  \textbf{8}(1) (2024). \doi{10.1145/3643504}

\bibitem{waft:2026}
Wang, Y., Deng, J.: {WAFT}: {W}arping-alone field transforms for optical flow.
  In: International Conference on Learning Representations (2026)

\bibitem{SEARAFT}
Wang, Y., Lipson, L., Deng, J.: {SEA-RAFT}: {S}imple, efficient, accurate
  {RAFT} for optical flow. In: European Conference on Computer Vision. vol.
  15065, pp. 36--54 (2024). \doi{10.1007/978-3-031-72667-5\_3}

\bibitem{witraj:2023}
Wu, D., Zeng, Y., Gao, R., Li, S., Li, Y., Shah, R.C., Lu, H., Zhang, D.:
  {WiTraj}: {R}obust indoor motion tracking with {WiFi} signals. IEEE
  Transactions on Mobile Computing  \textbf{22}(5),  3062--3078 (2023).
  \doi{10.1109/TMC.2021.3133114}

\bibitem{widir:2016}
Wu, D., Zhang, D., Xu, C., Wang, Y., Wang, H.: {WiDir}: Walking direction
  estimation using wireless signals. In: ACM International Joint Conference on
  Pervasive and Ubiquitous Computing. pp. 351--362 (2016).
  \doi{10.1145/2971648.2971658}

\bibitem{atheros-driver}
Xie, Y., Li, Z., Li, M.: Precise power delay profiling with commodity {WiFi}.
  In: ACM SIGMOBILE International Conference on Mobile Computing and
  Networking. p. 53–64 (2015). \doi{10.1145/2789168.2790124}

\bibitem{freesense:2018}
Xin, T., Guo, B., Wang, Z., Wang, P., Yu, Z.: {FreeSense}: {H}uman-behavior
  understanding using {Wi-Fi} signals. Journal of Ambient Intelligence and
  Humanized Computing  \textbf{9}(5),  1611--1622 (2018).
  \doi{10.1007/s12652-018-0793-4}

\bibitem{gmflow}
Xu, H., Zhang, J., Cai, J., Rezatofighi, H., Tao, D.: {GMFlow}: {L}earning
  optical flow via global matching. In: IEEE/CVF Conference on Computer Vision
  and Pattern Recognition. pp. 8121--8130 (2022).
  \doi{10.1109/CVPR52688.2022.00795}

\bibitem{Xu:2019:DFG}
Xu, R., Li, X., Zhou, B., Loy, C.C.: Deep flow-guided video inpainting. In:
  IEEE/CVF Conference on Computer Vision and Pattern Recognition. pp.
  3723--3732 (2019). \doi{10.1109/CVPR.2019.00384}

\bibitem{wistep:2018}
Xu, Y., Yang, W., Wang, J., Zhou, X., Li, H., Huang, L.: {WiStep}:
  {D}evice-free step counting with {WiFi} signals. ACM Interactive, Mobile,
  Wearable, and Ubiquitous Technologies  \textbf{1}(4) (2018).
  \doi{10.1145/3161415}

\bibitem{person-in-wifi-3d:2024}
Yan, K., Wang, F., Qian, B., Ding, H., Han, J., Wei, X.: {Person-in-WiFi 3D}:
  {E}nd-to-end multi-person {3D} pose estimation with {Wi-Fi}. In: IEEE/CVF
  Conference on Computer Vision and Pattern Recognition. pp. 969--978 (2024).
  \doi{10.1109/CVPR52733.2024.00098}

\bibitem{mmfi2023}
Yang, J., Huang, H., Zhou, Y., Chen, X., Xu, Y., Yuan, S., Zou, H., Lu, C.X.,
  Xie, L.: {MM-Fi}: {M}ulti-modal non-intrusive {4D} human dataset for
  versatile wireless sensing. In: Neural Information Processing Systems.
  vol.~36, pp. 18756--18768 (2023)

\bibitem{metafi:2022}
Yang, J., Zhou, Y., Huang, H., Zou, H., Xie, L.: {MetaFi}: {D}evice-free pose
  estimation via commodity {WiFi} for metaverse avatar simulation. In: IEEE
  World Forum on Internet of Things. pp.~1--6 (2022).
  \doi{10.1109/WF-IoT54382.2022.10152057}

\bibitem{FOLT:2023}
Yao, M., Wang, J., Peng, J., Chi, M., Liu, C.: {FOLT}: {F}ast multiple object
  tracking from {UAV}-captured videos based on optical flow. In: ACM
  International Conference on Multimedia. p. 3375–3383 (2023).
  \doi{10.1145/3581783.3611868}

\bibitem{wispeed:2018}
Zhang, F., Chen, C., Wang, B., Liu, K.J.R.: {WiSpeed}: {A} statistical
  electromagnetic approach for device-free indoor speed estimation. IEEE
  Internet of Things Journal  \textbf{5}(3),  2163--2177 (2018).
  \doi{10.1109/JIOT.2018.2826227}

\bibitem{Zhang:2023:EMA}
Zhang, G., Zhu, Y., Wang, H., Chen, Y., Wu, G., Wang, L.: Extracting motion and
  appearance via inter-frame attention for efficient video frame interpolation.
  In: IEEE/CVF Conference on Computer Vision and Pattern Recognition. pp.
  5682--5692 (2023). \doi{10.1109/CVPR52729.2023.00550}

\bibitem{WiDar3}
Zheng, Y., Zhang, Y., Qian, K., Zhang, G., Liu, Y., Wu, C., Yang, Z.:
  Zero-effort cross-domain gesture recognition with {Wi-Fi}. In: International
  Conference on Mobile Systems, Applications, and Services. p. 313–325
  (2019). \doi{10.1145/3307334.3326081}

\bibitem{Zhou:2023:PIP}
Zhou, S., Li, C., Chan, K.C.K., Loy, C.C.: {ProPainter}: {I}mproving
  propagation and transformer for video inpainting. In: IEEE/CVF International
  Conference on Computer Vision. pp. 10443--10452 (2023).
  \doi{10.1109/ICCV51070.2023.00961}

\bibitem{Zhu:2018:HTS}
Zhu, Y., Lan, Z., Newsam, S.D., Hauptmann, A.G.: Hidden two-stream
  convolutional networks for action recognition. In: Asian Conference on
  Computer Vision. vol. 11363, pp. 363--378 (2018).
  \doi{10.1007/978-3-030-20893-6\_23}

\bibitem{wifree:2018}
Zou, H., Zhou, Y., Yang, J., Spanos, C.J.: Device-free occupancy detection and
  crowd counting in smart buildings with {WiFi}-enabled {IoT}. Energy and
  Buildings  \textbf{174},  309--322 (2018).
  \doi{10.1016/j.enbuild.2018.06.040}

\end{thebibliography}


\begin{thebibliography}{99}
        \bibitem[95]{Jahedi:2024:HRM} Jahedi, A., Luz, M., Rivinius, M., Mehl, L., Bruhn, A.: MS-RAFT+: High resolution multi-scale RAFT. In: International Journal of Computer Vision \textbf{132}, pp 1835--1856 (2024). \doi{10.1007/s11263-023-01930-7}
        
        \bibitem[96]{Morimitsu:2021:plo} Morimitsu, H.: PyTorch lightning optical flow. \url{https://github.com/hmorimitsu/ptlflow} 
        (2021)
        
        \bibitem[97]{Morimitsu:2024:rpk} Morimitsu, H., Zhu, X., Ji, X., Yin, X.: Recurrent partial kernel network for efficient optical flow estimation. In: AAAI Conference on Artificial Intelligence. pp 4278--4286 (2024). \doi{10.1609/aaai.v38i5.28224}
        
        \bibitem[98]{Schmalfuss:2025:BRI} Schmalfuss, J., Oei, V., Mehl, L., Bartsch, M., Agnihotri, S., Keuper, M., Bruhn, A.: RobustSpring: Benchmarking robustness to image corruptions for optical flow, scene flow and stereo. In: CoRR abs/2505.09368 (2025). \doi{10.48550/ARXIV.2505.09368}
        
        \bibitem[99]{Zheng:2020:OFD} Zheng, Y., Zhang, M., Lu, F.: Optical flow in the dark. In: IEEE/CVF Conference on Computer Vision and Pattern Recognition. pp 6748--6756 (2020). \doi{10.1109/CVPR42600.2020.00678}
        
    \end{thebibliography}

\clearpage
\refstepcounter{chapter}%
\appendix
\renewcommand{\thepage}{\roman{page}}
\setcounter{page}{1}
\renewcommand{\thefigure}{A.\arabic{figure}}
\setcounter{figure}{0}
\renewcommand{\thetable}{A.\arabic{table}} 
\setcounter{table}{0}
\renewcommand{\theequation}{A.\arabic{equation}} 
\setcounter{equation}{0}

\maketitlesupplementary
\phantomsection
\label{app:supplement-master}
\FloatBarrier

\section{WiFlow Dataset}
\label{suppl:action-details}

\begin{table}[hbt]
    \centering
    \caption{\textbf{Action definitions.} Description of each recorded action that we performed in our capturing room. All actions were performed while walking, some included faster motion or brief periods of stationary motion.}
    
    \small
    \begin{tabularx}{\linewidth}{>{\hsize=0.3\hsize}X
    >{\hsize=1.7\hsize}X}
    \toprule
    \textbf{Action} & \textbf{Description} \\
    \midrule
    Slow-walk & The subject walks continuously at a regular speed, producing longer stance phases and smoother transitions. \\

    Fast-walk & The subject walks continuously at an increased speed compared to normal gait, with no additional upper-body activity. \\

    Together & Slow-walk with two subjects, with unequal motions in terms of positions and directions.\\
    
    Collect & The subject walks through the movement area while bending intermittently to pick up imaginary objects. The motion alternates between forward walking and short stationary collection phases. \\
    
    Knee & The subject transitions from walking to a kneeling posture on the floor and remains briefly stationary, then stands up and continues walking. \\
    
    Mill & The subject walks continuously while performing a single arm rotation. \\

    Waving & The subject walks while repeatedly waving one arm, introducing periodic upper-body motion with brief quasi-stationary arm phases. \\
    \midrule
    Off-area &  No subject is present in the movement area, but there is movement in the adjacent off-area in the same room.\\
        
    Void & No subject is present in the movement area, and there is no intended motion in the off-area.\\

    \bottomrule
    \end{tabularx}
    \label{tab:actions}
\end{table}

\begin{table}[hbt]
    \centering
    \caption{\textbf{Dataset overview.} Our dataset consists of $7$ different actions performed multiple times by a total of $10$ people and $2$ additional actions where no one is moving (\void) and one where there is only motion outside of the caption area (\noise). We capture camera frames and CSI and derive three datasets (\sideview, \birdview, \birdviewplus).}
    
    \setlength{\tabcolsep}{0.1cm}%
    \footnotesize%

    \begin{tabularx}{\colwidth}{X X S[table-format=2,group-separator={,},group-minimum-digits=4] S[table-format=5,group-separator={,},group-minimum-digits=4]@{\hspace{0.1cm}} S[table-format=5,group-separator={,},group-minimum-digits=4] S[table-format=6,group-separator={,},group-minimum-digits=4]@{\hspace{0.1cm}} S[table-format=6,group-separator={,},group-minimum-digits=4]@{\hspace{0.1cm}} S[table-format=7,group-separator={,},group-minimum-digits=4]}
        \toprule 
        \multirow{2}{*}[-0.5\dimexpr \aboverulesep + \belowrulesep + \cmidrulewidth]{Action} &
        \multirow{2}{*}[-0.5\dimexpr \aboverulesep + \belowrulesep + \cmidrulewidth]{ID} &
        \multicolumn{1}{c}{\multirow{2}{*}[-0.5\dimexpr \aboverulesep + \belowrulesep + \cmidrulewidth]{Subjects}} & 
        \multicolumn{2}{c}{Camera Frames} & 
        \multicolumn{3}{c}{CSI}\\
        \cmidrule(lr){4-5}
        \cmidrule(lr){6-8}
         & & \multicolumn{1}{c}{} & {\sideview} & {\birdview} & {\sideview} & {\birdview} & {\birdviewplus}\\
        \midrule
        Slow-walk       & A21--A28 & 10 & 24000 & 14400 & 240000 & 144000 & 1440000\\
        Fast-walk       & A17--A20 & 10 & 11700 & 7020 & 117000 & 70200 & 702000\\
        Together        & A13--A16 & 10 & 12000 & 7200 & 120000 & 72000 & 720000\\
        Collect         & A01--A04 & 10 & 12000 & 7200 & 120000 & 72000 & 720000\\
        Knee            & A05--A08 & 10 & 11780 & 7068 & 117800 & 70680 & 706800\\
        Mill            & A09--A12 & 10 & 12000 & 7200 & 120000 & 72000 & 720000\\
        Waving          & A29--A32 & 9  & 10800 & 6480 & 108000 & 64800 & 648000\\
        Off-area        & A33--A36 & 0 & 1200 & 720 & 12000 & 7200 & 72000\\
        Void            & A37--A44 & 0 & 2400 & 1440 & 24000 & 14400 & 144000\\
        \midrule
        Total & & \multicolumn{1}{c}{} & 97880 & 58728 & 978800 & 587280 & 5872800 \\
        \bottomrule
    \end{tabularx}

    \label{tab:seemodataset}
\end{table}

\begin{table}[hbt]
    \centering
    \caption{\textbf{Split protocol} details showing the sequences and subjects used as training, validation, and testing set in our \textit{time} and \textit{subject split}.}

\begin{tabularx}{\linewidth}{@{} >{\centering\arraybackslash}l >{\centering\arraybackslash}l >{\centering\arraybackslash}X >{\centering\arraybackslash}X >{\centering\arraybackslash}X @{}}
\toprule
\textbf{Protocol} & \textbf{Attribute} & \textbf{Train} & \textbf{Validation} & \textbf{Test} \\
\midrule
\multirow{3}{*}{\footnotesize \textit{Time}}
 & \footnotesize Actions  & A02--A04, A06--A08, A10--A12, A14--A16, A18--A20, A22--A24, A26--A28, A30--A32, A34--A36, A38--A40, A42--A44 & A25 & A01, A05, A09, A13, A17, A21, A29, A33, A37, A41 \\
 & \footnotesize Subjects & S00--S10 & S00--S10 & S00--S10 \\
 & \footnotesize Percentage & 75\% & 3\% & 22\% \\
\addlinespace[0.5em]
\midrule
\addlinespace[0.3em]
\multirow{3}{*}{\footnotesize \textit{Subject}}
 & \footnotesize Actions  & A01--A44 & A01--A32 & A01--A32 \\
 & \footnotesize Subjects & S00, S01, S03--S05, S07, S09, S10 & S02 & S06, S08 \\
 & \footnotesize Percentage & 71\% & 10\% & 19\% \\
\bottomrule
\end{tabularx}

    \label{tab:split-detail}
\end{table}

Our dataset consists of $9$ different actions, shortly described in \cref{tab:actions}. We recorded up to $10$ persons (labeled S01-S10) performing multiple repetitions of each action, resulting in multiple sequences denoted as A01--A44. Additionally, we introduce a pseudo-subject S00 for the two special actions \textit{off-area} and \textit{void}. A summary of the total number of captured camera frames and channel state information (CSI) is provided in \cref{tab:seemodataset}.

For our two proposed split protocols for training and evaluation, we provide the distribution \wrt sequence and subject numbers in \cref{tab:split-detail}. Our \textit{time split} allocates $75\%$ of our captured sequences to training, $3\%$ to validation, and the remaining $22\%$ to testing. In contrast, our \textit{subject split} uses all captures of $7$ subjects for training, one subject for validation, and two for testing.

\section{Generation of Pseudo Ground Truth }
\label{suppl:pseudo-detail}
Generating the pseudo ground truths (\pseudo) from our captured camera frames is an important step in the construction of our dataset. Before processing, we downsample the video frames to $168 \times 128$ pixels. To ensure reliable optical flow estimations for our \pseudo, we create an ensemble of multiple optical flow methods and average their predictions. We provide details on the optical flow methods and checkpoints used in \cref{tab:flowmodels}. Finally, to minimize background noise, we threshold all motion with a magnitude less than $0.5$ to zero.

\begin{table}
\centering
\caption{\textbf{Optical flow model} names and their respective checkpoints within the PTLFlow Framework~\cite{Morimitsu:2021:plo} that we utilized in our ensemble to create \pseudo optical flow predictions for our proposed datasets.}
\label{tab:flowmodels}
\begin{tabularx}{\linewidth}{>{\centering\arraybackslash}X >{\centering\arraybackslash}X}
\toprule
\textbf{Model Name} & \textbf{Fine-tuned Dataset Tag} \\
\midrule
rpknet~\cite{Morimitsu:2024:rpk} & things \\
ms\_raft\_p~\cite{Jahedi:2024:HRM} & mixed \\
sea\_raft\_m~\cite{SEARAFT} & spring \\
memflow~\cite{Memflow2024} & spring \\
dpflow~\cite{DPFlow2025} & spring \\
\bottomrule
\end{tabularx}
\end{table}

\section{Architecture Details}
Both of our architectures, \ourm and \ourl, contain a mask block that only predicts where movements occur, rather than their direction and magnitude. For training these two architectures, we found that it is best to pre-train only the mask block using a mean squared error loss ($\mathcal{L}_{MSE}$) between the mask prediction and a \pseudo where all non-zero motions are set to $1$ before training the other modules of each architecture.

\section{Qualitative Results}
\label{app:qualitative}
In the supplemental files, we include a video showcasing further qualitative results for all our models across different split protocols and actions.

\section{Detailed Results}
\label{suppl:per-action}
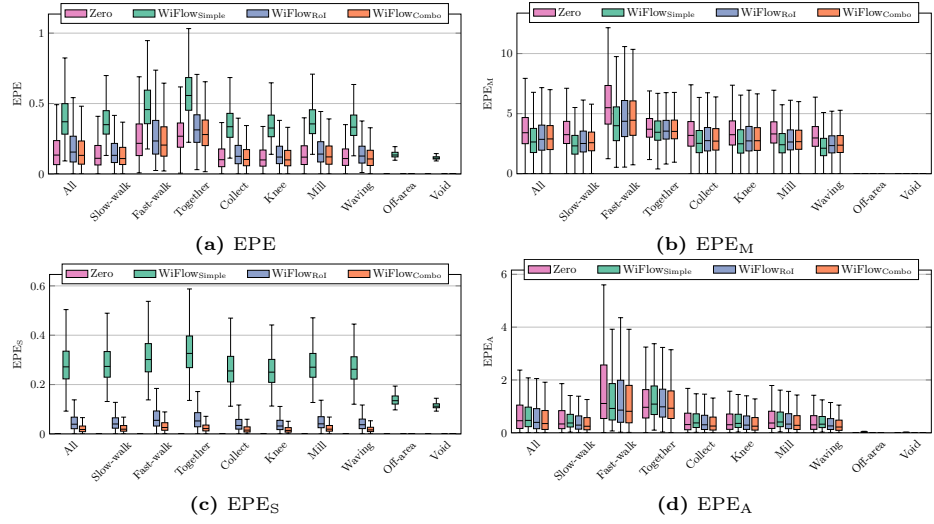
\begin{figure}[hbt]
    \centering

    \begin{subfigure}{0.49\linewidth}
        \centering
        \begin{tikzpicture}

    \pgfplotstableread[col sep=comma]{
        0.0,0.06610141135752201,0.13398142904043198,0.23679225891828537,0.49193015694618225
        0.0,0.06370361149311066,0.11112730577588081,0.20221934095025063,0.4093572497367859
        0.00770538067445159,0.13040828332304955,0.21728389710187912,0.35507412999868393,0.6900599598884583
        0.005506961140781641,0.18970485031604767,0.2678229957818985,0.36181218922138214,0.6179717183113098
        0.0,0.05158919282257557,0.1012667641043663,0.17740508541464806,0.36483800411224365
        0.0,0.055229091085493565,0.09954029694199562,0.169867891818285,0.3372052013874054
        0.0,0.06689119152724743,0.11939318478107452,0.20005451142787933,0.39865395426750183
        0.0,0.06068018451333046,0.10916905105113983,0.1775282770395279,0.3501783609390259
        0.0,0.0,0.0,0.0,0.0
        0.0,0.0,0.0,0.0,0.0
    }\zerodata
    \pgfplotstabletranspose\zerodatatransposed{\zerodata}
    
    \pgfplotstableread[col sep=comma]{
        0.09222942590713501,0.28245633840560913,0.37044060230255127,0.4990897700190544,0.8238505721092224
        0.13124321401119232,0.28294577449560165,0.34992237389087677,0.4493474066257477,0.6986904144287109
        0.1771944910287857,0.35679416358470917,0.45700548589229584,0.5949981957674026,0.9468938112258911
        0.22465428709983826,0.4522877708077431,0.5571486353874207,0.6844045668840408,1.0322757959365845
        0.11214097589254379,0.26006339490413666,0.33538496494293213,0.4301532283425331,0.6844947338104248
        0.13931772112846375,0.26428715884685516,0.3260316997766495,0.4178242087364197,0.6471377611160278
        0.1393229365348816,0.2863667160272598,0.35533374547958374,0.45576243847608566,0.7086887359619141
        0.12865599989891052,0.273377388715744,0.33242347836494446,0.41826340556144714,0.6353000402450562
        0.09726567566394806,0.12090787291526794,0.1345444917678833,0.1537410467863083,0.19382554292678833
        0.09222942590713501,0.10603877529501915,0.11220665276050568,0.12198769301176071,0.1445043981075287
    }\simpledata
    \pgfplotstabletranspose\simpledatatransposed{\simpledata}
    
    \pgfplotstableread[col sep=comma]{
        0.0,0.0849379412829876,0.15532495826482773,0.2679015100002289,0.5418419241905212
        0.0,0.08094486966729164,0.13207103312015533,0.2167489156126976,0.41580578684806824
        0.02438499592244625,0.14100683480501175,0.23406774550676346,0.3800986781716347,0.7371663451194763
        0.030319316312670708,0.22443712502717972,0.312985897064209,0.4201318025588989,0.706926703453064
        0.0,0.07272505201399326,0.12493840605020523,0.20271410420536995,0.39617276191711426
        0.0,0.0732407458126545,0.11994015797972679,0.19686388969421387,0.38035550713539124
        0.0,0.08501730859279633,0.14106109738349915,0.22895538806915283,0.4434650242328644
        0.009908787906169891,0.07718480378389359,0.12702830135822296,0.19701676070690155,0.37620189785957336
        0.0,0.0,0.0,0.0,0.0
        0.0,0.0,0.0,0.0,0.0
    }\roidata
    \pgfplotstabletranspose\roidatatransposed{\roidata}
    
    \pgfplotstableread[col sep=comma]{
        1.8556829672888853e-06,0.0693670678883791,0.1318896934390068,0.23408342525362968,0.4809468686580658
        0.00033024942968040705,0.06681410782039165,0.10925982147455215,0.18876510858535767,0.3677307069301605
        0.02093140222132206,0.12610559165477753,0.20494700968265533,0.3348424807190895,0.6448287963867188
        0.015751831233501434,0.20134077221155167,0.2797120362520218,0.38287919759750366,0.6547860503196716
        1.600064024387393e-05,0.05817124526947737,0.10252406820654869,0.17298226431012154,0.34358662366867065
        2.99448238365585e-05,0.05778568610548973,0.09967987611889839,0.16727090999484062,0.3310054540634155
        7.96465392340906e-05,0.06894143112003803,0.1209372878074646,0.19842329248785973,0.3908662796020508
        0.0011529292678460479,0.059982091188430786,0.10640482604503632,0.1681349128484726,0.32799309492111206
        1.988779104067362e-06,5.153206075192429e-06,9.305794264946599e-06,2.0246177882654592e-05,3.914121771231294e-05
        1.8556829672888853e-06,3.517711263612e-06,4.4096973397245165e-06,5.848774435435189e-06,9.06434615899343e-06
    }\combodata
    \pgfplotstabletranspose\combodatatransposed{\combodata}

    \pgfplotstableread[header=true]{
        Name
All
Slow-walk
Fast-walk
Together
Collect
Knee
Mill
Waving
Off-area
Void
    }\namesdata

    \begin{axis}[
        boxplot/draw direction = y,
        boxplot/box extend=0.15,
        ymajorgrids,
        xtick = {1, 2, 3, 4, 5, 6, 7, 8, 9, 10},
        xmin = 0.5,
        xmax = 10.5,
        ymin = 0,
        xticklabel style = {align=center, font=\small, rotate=45},
        xticklabels from table={\namesdata}{Name},
        table/col sep=comma,
        xtick style = {draw=none},
        ylabel = {EPE},
        cycle list/Pastel2,
        width=130mm,
        height=60mm,
        boxplot/every median/.style={
            draw=black,
            postaction={
                decorate,
                decoration={
                    markings,
                    mark=at position 0.5 with {
                        \node[above, font=\tiny, yshift=-1.5pt, color=black] {
                        };
                    }
                }
            }
        },
        extra description code/.append style={/pgfplots/table/col sep=comma},
        legend style={
            at={(0.5,0.94)},
            anchor=south, 
            legend columns=4,
            font=\footnotesize,
            /tikz/every even column/.append style={column sep=0.5cm}
        }
    ]
            \foreach \n in {1,...,10} {
            \addplot+[boxplot, fill, draw=black, boxplot/draw position={\n-0.3}, fill=zero, forget plot] table[y index=\n] {\zerodatatransposed};
        }
        \addlegendimage{fill=zero, draw=black, area legend}
        \addlegendentry{Zero}
        
        \foreach \n in {1,...,10} {
            \addplot+[boxplot, fill, draw=black, boxplot/draw position={\n-0.1}, fill=simple, forget plot] table[y index=\n] {\simpledatatransposed};
        }
        \addlegendimage{fill=simple, draw=black, area legend}
        \addlegendentry{\ours}
        
        \foreach \n in {1,...,10} {
            \addplot+[boxplot, fill, draw=black, boxplot/draw position={\n+0.1}, fill=roi, forget plot] table[y index=\n] {\roidatatransposed};
        }
        \addlegendimage{fill=roi, draw=black, area legend}
        \addlegendentry{\ourm}
        
        \foreach \n in {1,...,10} {
            \addplot+[boxplot, fill, draw=black, boxplot/draw position={\n+0.3}, fill=combo, forget plot] table[y index=\n] {\combodatatransposed};
        }
        \addlegendimage{fill=combo, draw=black, area legend}
        \addlegendentry{\ourl}
    \end{axis}
\end{tikzpicture}
        \caption{EPE}
    \end{subfigure}
    \hfill
    \begin{subfigure}{0.49\linewidth}
        \centering
        \begin{tikzpicture}
    \usetikzlibrary{decorations.markings}

    \pgfplotstableread[col sep=comma]{
        0.0,2.498010277748108,3.410646915435791,4.674451470375061,7.936241626739502
        0.0,2.5128304958343506,3.2502647638320923,4.355103850364685,7.1116251945495605
        1.2458384037017822,4.1300084590911865,5.494056701660156,7.342258810997009,12.151994705200195
        1.1761889457702637,3.0416939854621887,3.7031995058059692,4.594577789306641,6.888514518737793
        0.0,2.2928101420402527,3.1876333951950073,4.338420867919922,7.400989532470703
        0.0,2.3885075449943542,3.2458654642105103,4.395182251930237,7.367001533508301
        0.0,2.552697241306305,3.3104649782180786,4.319167971611023,6.9490180015563965
        0.0,2.2799675464630127,2.9740636348724365,3.9232356548309326,6.373498916625977
        0.0,0.0,0.0,0.0,0.0
        0.0,0.0,0.0,0.0,0.0
    }\zerodata
    \pgfplotstabletranspose\zerodatatransposed{\zerodata}
    
    \pgfplotstableread[col sep=comma]{
        0.0,1.7614257335662842,2.6408878564834595,3.769225597381592,6.773714542388916
        0.0,1.6287067234516144,2.3144233226776123,3.1880266070365906,5.510842800140381
        0.5319308638572693,2.7513538002967834,3.994408965110779,5.560962677001953,9.747114181518555
        0.3940448462963104,2.7850736379623413,3.450469732284546,4.390944242477417,6.689171314239502
        0.0,1.7520109713077545,2.527021050453186,3.59607470035553,6.349467754364014
        0.0,1.714658498764038,2.4897913932800293,3.6608930230140686,6.542783260345459
        0.0,1.7375313639640808,2.411444067955017,3.3408557772636414,5.739622116088867
        0.0,1.5012811422348022,2.120626449584961,2.9428138732910156,5.0926971435546875
        0.0,0.0,0.0,0.0,0.0
        0.0,0.0,0.0,0.0,0.0
    }\simpledata
    \pgfplotstabletranspose\simpledatatransposed{\simpledata}
    
    \pgfplotstableread[col sep=comma]{
        0.0,1.958124577999115,2.859746217727661,4.043253779411316,7.163879871368408
        0.0,1.8109334409236908,2.5146241188049316,3.5460525155067444,6.132408142089844
        0.5480102300643921,3.0897085666656494,4.36442494392395,6.091917157173157,10.582157135009766
        0.8040198683738708,2.8817050457000732,3.5428075790405273,4.44029700756073,6.74669075012207
        0.0,1.8971113562583923,2.751721978187561,3.8437663316726685,6.736155986785889
        0.0,1.8964348435401917,2.7319648265838623,3.9338698983192444,6.951312065124512
        0.0,1.9840284287929535,2.639259696006775,3.6459136605262756,6.114810943603516
        0.0,1.7329754829406738,2.337303876876831,3.154311180114746,5.203483581542969
        0.0,0.0,0.0,0.0,0.0
        0.0,0.0,0.0,0.0,0.0
    }\roidata
    \pgfplotstabletranspose\roidatatransposed{\roidata}
    
    \pgfplotstableread[col sep=comma]{
        0.0,2.008876621723175,2.894320845603943,4.00529408454895,6.995463848114014
        0.0,1.9091147482395172,2.580709934234619,3.463632583618164,5.787764072418213
        0.7389782071113586,3.193512499332428,4.4499993324279785,6.060792803764343,10.350951194763184
        0.95705246925354,2.9063215851783752,3.5203784704208374,4.461239695549011,6.766445159912109
        0.0,1.9647980034351349,2.7122966051101685,3.752740442752838,6.392439842224121
        0.0,1.9381189346313477,2.7448326349258423,3.8332876563072205,6.647187232971191
        0.0,2.029223918914795,2.6656227111816406,3.620619833469391,6.000345706939697
        0.0,1.7652084827423096,2.3739840984344482,3.175900459289551,5.2798967361450195
        0.0,0.0,0.0,0.0,0.0
        0.0,0.0,0.0,0.0,0.0
    }\combodata
    \pgfplotstabletranspose\combodatatransposed{\combodata}

    \pgfplotstableread[header=true]{
        Name
All
Slow-walk
Fast-walk
Together
Collect
Knee
Mill
Waving
Off-area
Void
    }\namesdata

    \begin{axis}[
        boxplot/draw direction = y,
        boxplot/box extend=0.15,
        ymajorgrids,
        xtick = {1, 2, 3, 4, 5, 6, 7, 8, 9, 10},
        xmin = 0.5,
        xmax = 10.5,
        ymin = 0,
        xticklabel style = {align=center, font=\small, rotate=45},
        xticklabels from table={\namesdata}{Name},
        table/col sep=comma,
        xtick style = {draw=none},
        ylabel = {EPE$_\text{M}$},
        cycle list/Pastel2,
        width=130mm,
        height=60mm,
        boxplot/every median/.style={
            draw=black,
            postaction={
                decorate,
                decoration={
                    markings,
                    mark=at position 0.5 with {
                        \node[above, font=\tiny, yshift=-1.5pt, color=black] {
                        };
                    }
                }
            }
        },
        extra description code/.append style={/pgfplots/table/col sep=comma},
        legend style={
            at={(0.5,0.94)},
            anchor=south, 
            legend columns=4,
            font=\footnotesize,
            /tikz/every even column/.append style={column sep=0.5cm}
        }
    ]
            \foreach \n in {1,...,10} {
            \addplot+[boxplot, fill, draw=black, boxplot/draw position={\n-0.3}, fill=zero, forget plot] table[y index=\n] {\zerodatatransposed};
        }
        \addlegendimage{fill=zero, draw=black, area legend}
        \addlegendentry{Zero}
        
        \foreach \n in {1,...,10} {
            \addplot+[boxplot, fill, draw=black, boxplot/draw position={\n-0.1}, fill=simple, forget plot] table[y index=\n] {\simpledatatransposed};
        }
        \addlegendimage{fill=simple, draw=black, area legend}
        \addlegendentry{\ours}
        
        \foreach \n in {1,...,10} {
            \addplot+[boxplot, fill, draw=black, boxplot/draw position={\n+0.1}, fill=roi, forget plot] table[y index=\n] {\roidatatransposed};
        }
        \addlegendimage{fill=roi, draw=black, area legend}
        \addlegendentry{\ourm}
        
        \foreach \n in {1,...,10} {
            \addplot+[boxplot, fill, draw=black, boxplot/draw position={\n+0.3}, fill=combo, forget plot] table[y index=\n] {\combodatatransposed};
        }
        \addlegendimage{fill=combo, draw=black, area legend}
        \addlegendentry{\ourl}
    \end{axis}
\end{tikzpicture}
        \caption{EPE$_\text{M}$}
    \end{subfigure}

    \vfill

    \begin{subfigure}{0.49\linewidth}
        \centering
        \begin{tikzpicture}
    \usetikzlibrary{decorations.markings}

        \pgfplotstableread[col sep=comma]{
        0.0,0.0,0.0,0.0,0.0
        0.0,0.0,0.0,0.0,0.0
        0.0,0.0,0.0,0.0,0.0
        0.0,0.0,0.0,0.0,0.0
        0.0,0.0,0.0,0.0,0.0
        0.0,0.0,0.0,0.0,0.0
        0.0,0.0,0.0,0.0,0.0
        0.0,0.0,0.0,0.0,0.0
        0.0,0.0,0.0,0.0,0.0
        0.0,0.0,0.0,0.0,0.0
    }\zerodata
    \pgfplotstabletranspose\zerodatatransposed{\zerodata}
    
    \pgfplotstableread[col sep=comma]{
        0.09222942590713501,0.22295886650681496,0.2717657834291458,0.33532722294330597,0.5038270354270935
        0.13124321401119232,0.2298525609076023,0.2735027074813843,0.3337225541472435,0.4892858564853668
        0.1377977430820465,0.25161632150411606,0.30142639577388763,0.3659268468618393,0.537367045879364
        0.13560084998607635,0.26857515424489975,0.3262331932783127,0.3968384116888046,0.5875211954116821
        0.11214097589254379,0.2106373868882656,0.2551117539405823,0.31429798156023026,0.4693430960178375
        0.1127873957157135,0.2088874839246273,0.2502811551094055,0.3019817918539047,0.44158223271369934
        0.12728692591190338,0.22959114983677864,0.2703777402639389,0.3261484056711197,0.47060006856918335
        0.120262511074543,0.22206617891788483,0.26196736097335815,0.3126536011695862,0.44523295760154724
        0.09726567566394806,0.12090787291526794,0.1345444917678833,0.1537410467863083,0.19382554292678833
        0.09222942590713501,0.10603877529501915,0.11220665276050568,0.12198769301176071,0.1445043981075287
    }\simpledata
    \pgfplotstabletranspose\simpledatatransposed{\simpledata}
    
    \pgfplotstableread[col sep=comma]{
        0.0,0.021025685127824545,0.03923744149506092,0.06784870103001595,0.1380559206008911
        0.0,0.02309094602242112,0.040212513878941536,0.06517508625984192,0.12778763473033905
        0.0,0.03183356486260891,0.05504545755684376,0.09290456026792526,0.18423455953598022
        0.0,0.02884636865928769,0.052452268078923225,0.08607742376625538,0.17158867418766022
        0.0,0.01934510050341487,0.034143779426813126,0.05864954087883234,0.11716528236865997
        0.0,0.017434855923056602,0.0320086944848299,0.0550451148301363,0.11127699911594391
        0.0,0.025062732864171267,0.0407754871994257,0.06972445920109749,0.1365625113248825
        0.0,0.021437492221593857,0.03719916567206383,0.060055989772081375,0.11713223904371262
        0.0,0.0,0.0,0.0,0.0
        0.0,0.0,0.0,0.0,0.0
    }\roidata
    \pgfplotstabletranspose\roidatatransposed{\roidata}
    
    \pgfplotstableread[col sep=comma]{
        1.8556829672888853e-06,0.009072462096810341,0.017654909752309322,0.03184077609330416,0.06598875671625137
        0.00013882671191822737,0.010964836459606886,0.019033164717257023,0.03377074282616377,0.06791555136442184
        0.0009358328534290195,0.014710047049447894,0.025875712744891644,0.04535788297653198,0.08910377323627472
        9.965422941604629e-05,0.012101363157853484,0.02155214175581932,0.03579636011272669,0.07125511765480042
        1.600064024387393e-05,0.007666231016628444,0.015279179904609919,0.028509192634373903,0.05941682681441307
        6.824840966146439e-06,0.007129814242944121,0.014071593061089516,0.02548717288300395,0.051211778074502945
        7.96465392340906e-05,0.01105038681998849,0.01922935340553522,0.034039164893329144,0.06825863569974899
        4.0198367059929296e-05,0.009311871603131294,0.016174590215086937,0.026872543618083,0.05316817760467529
        1.988779104067362e-06,5.153206075192429e-06,9.305794264946599e-06,2.0246177882654592e-05,3.914121771231294e-05
        1.8556829672888853e-06,3.517711263612e-06,4.4096973397245165e-06,5.848774435435189e-06,9.06434615899343e-06
    }\combodata
    \pgfplotstabletranspose\combodatatransposed{\combodata}

    \pgfplotstableread[header=true]{
        Name
All
Slow-walk
Fast-walk
Together
Collect
Knee
Mill
Waving
Off-area
Void
    }\namesdata

    \begin{axis}[
        boxplot/draw direction = y,
        boxplot/box extend=0.15,
        ymajorgrids,
        xtick = {1, 2, 3, 4, 5, 6, 7, 8, 9, 10},
        xmin = 0.5,
        xmax = 10.5,
        ymin = 0,
        xticklabel style = {align=center, font=\small, rotate=45},
        xticklabels from table={\namesdata}{Name},
        table/col sep=comma,
        xtick style = {draw=none},
        ylabel = {EPE$_\text{S}$},
        cycle list/Pastel2,
        width=130mm,
        height=60mm,
        boxplot/every median/.style={
            draw=black,
            postaction={
                decorate,
                decoration={
                    markings,
                    mark=at position 0.5 with {
                        \node[above, font=\tiny, yshift=-1.5pt, color=black] {
                        };
                    }
                }
            }
        },
        extra description code/.append style={/pgfplots/table/col sep=comma},
        legend style={
            at={(0.5,0.94)},
            anchor=south, 
            legend columns=4,
            font=\footnotesize,
            /tikz/every even column/.append style={column sep=0.5cm}
        }
    ]
            \foreach \n in {1,...,10} {
            \addplot+[boxplot, fill, draw=black, boxplot/draw position={\n-0.3}, fill=zero, forget plot] table[y index=\n] {\zerodatatransposed};
        }
        \addlegendimage{fill=zero, draw=black, area legend}
        \addlegendentry{Zero}
        
        \foreach \n in {1,...,10} {
            \addplot+[boxplot, fill, draw=black, boxplot/draw position={\n-0.1}, fill=simple, forget plot] table[y index=\n] {\simpledatatransposed};
        }
        \addlegendimage{fill=simple, draw=black, area legend}
        \addlegendentry{\ours}
        
        \foreach \n in {1,...,10} {
            \addplot+[boxplot, fill, draw=black, boxplot/draw position={\n+0.1}, fill=roi, forget plot] table[y index=\n] {\roidatatransposed};
        }
        \addlegendimage{fill=roi, draw=black, area legend}
        \addlegendentry{\ourm}
        
        \foreach \n in {1,...,10} {
            \addplot+[boxplot, fill, draw=black, boxplot/draw position={\n+0.3}, fill=combo, forget plot] table[y index=\n] {\combodatatransposed};
        }
        \addlegendimage{fill=combo, draw=black, area legend}
        \addlegendentry{\ourl}
    \end{axis}
\end{tikzpicture}
        \caption{EPE$_\text{S}$}
    \end{subfigure}
    \hfill
    \begin{subfigure}{0.49\linewidth}
        \centering
        \begin{tikzpicture}
    \usetikzlibrary{decorations.markings}

    \pgfplotstableread[col sep=comma]{
        0.0,0.1648590974509716,0.45663467049598694,1.0487474203109741,2.3742406368255615
        0.0,0.1567087173461914,0.33469559252262115,0.8386352509260178,1.8609848022460938
        0.009052763693034649,0.5400658249855042,1.1096509099006653,2.567280888557434,5.600358009338379
        0.006835108157247305,0.5604478865861893,0.9658355116844177,1.6352749466896057,3.246244192123413
        0.0,0.1152182836085558,0.3110169768333435,0.7423687726259232,1.681004524230957
        0.0,0.13012216240167618,0.3052264004945755,0.7113178372383118,1.5745450258255005
        0.0,0.16479134932160378,0.3703855574131012,0.816690057516098,1.79033625125885
        0.0,0.13097071647644043,0.29997891187667847,0.6486697793006897,1.4213651418685913
        0.0,0.0,0.0,0.0,0.0
        0.0,0.0,0.0,0.0,0.0
    }\zerodata
    \pgfplotstabletranspose\zerodatatransposed{\zerodata}
    
    \pgfplotstableread[col sep=comma]{
        0.010014921426773071,0.23203417286276817,0.4698522984981537,0.9716945588588715,2.0799953937530518
        0.023510927334427834,0.21968530490994453,0.37137190997600555,0.6984752863645554,1.4114009141921997
        0.06848454475402832,0.4844832196831703,0.9218769967556,1.8631776869297028,3.9206173419952393
        0.10009750723838806,0.6913394182920456,1.0862218737602234,1.772910714149475,3.3725249767303467
        0.015993725508451462,0.19988729432225227,0.37375448644161224,0.7178990542888641,1.4778838157653809
        0.02795751951634884,0.20236628875136375,0.35599128901958466,0.7023084610700607,1.4500024318695068
        0.03474026173353195,0.23141401261091232,0.40873146057128906,0.7856060415506363,1.6161364316940308
        0.023986995220184326,0.19666437804698944,0.3265781104564667,0.6170064210891724,1.247295618057251
        0.011421862989664078,0.016965026035904884,0.022950749844312668,0.034824222326278687,0.056247223168611526
        0.010014921426773071,0.012876564171165228,0.014661295339465141,0.017654614988714457,0.024035820737481117
    }\simpledata
    \pgfplotstabletranspose\simpledatatransposed{\simpledata}
    
    \pgfplotstableread[col sep=comma]{
        0.0,0.15457375720143318,0.39467889070510864,0.9143569022417068,2.0537962913513184
        0.0,0.13823939859867096,0.29131028056144714,0.6396037638187408,1.3858023881912231
        0.021088765934109688,0.3980792388319969,0.8584733009338379,1.9886791408061981,4.357589244842529
        0.028306733816862106,0.5956975668668747,0.9831976890563965,1.6509799361228943,3.232109546661377
        0.0,0.12420393712818623,0.30324430763721466,0.6626976877450943,1.4631367921829224
        0.0,0.12855437770485878,0.2914045751094818,0.6377275437116623,1.4003733396530151
        0.0,0.1554291695356369,0.32996895909309387,0.7231346666812897,1.5678210258483887
        0.002757071750238538,0.1249903067946434,0.26015087962150574,0.5365670919418335,1.146682620048523
        0.0,0.0,0.0,0.0,0.0
        0.0,0.0,0.0,0.0,0.0
    }\roidata
    \pgfplotstabletranspose\roidatatransposed{\roidata}
    
    \pgfplotstableread[col sep=comma]{
        2.5249035501273553e-11,0.1265123337507248,0.3491652309894562,0.8424738943576813,1.9147286415100098
        4.009193162346492e-06,0.11397876776754856,0.24731378257274628,0.5746576339006424,1.2593345642089844
        0.01395120844244957,0.37837647646665573,0.8137346804141998,1.7954705357551575,3.9182021617889404
        0.010503392666578293,0.5410237461328506,0.9296877086162567,1.586213380098343,3.1452088356018066
        1.918828207436718e-09,0.10363678447902203,0.2556409388780594,0.5948359966278076,1.3169862031936646
        1.794499127072413e-08,0.102187380194664,0.2501279413700104,0.5766726285219193,1.2836999893188477
        9.195336048151148e-08,0.12449167482554913,0.28306715190410614,0.6486577242612839,1.4316120147705078
        5.967255856376141e-05,0.09306402504444122,0.21814392507076263,0.47938457131385803,1.0497629642486572
        3.518264152235773e-11,2.3687249384174436e-10,6.898013271694481e-10,6.3089355961665206e-09,1.227143098248007e-08
        2.5249035501273553e-11,1.5945463005540006e-10,2.7032108496083396e-10,5.032215061318013e-10,1.0139685757692973e-09
    }\combodata
    \pgfplotstabletranspose\combodatatransposed{\combodata}

    \pgfplotstableread[header=true]{
        Name
All
Slow-walk
Fast-walk
Together
Collect
Knee
Mill
Waving
Off-area
Void
    }\namesdata

    \begin{axis}[
        boxplot/draw direction = y,
        boxplot/box extend=0.15,
        ymajorgrids,
        xtick = {1, 2, 3, 4, 5, 6, 7, 8, 9, 10},
        xmin = 0.5,
        xmax = 10.5,
        ymin = 0,
        xticklabel style = {align=center, font=\small, rotate=45},
        xticklabels from table={\namesdata}{Name},
        table/col sep=comma,
        xtick style = {draw=none},
        ylabel = {EPE$_\text{A}$},
        cycle list/Pastel2,
        width=130mm,
        height=60mm,
        boxplot/every median/.style={
            draw=black,
            postaction={
                decorate,
                decoration={
                    markings,
                    mark=at position 0.5 with {
                        \node[above, font=\tiny, yshift=-1.5pt, color=black] {
                        };
                    }
                }
            }
        },
        extra description code/.append style={/pgfplots/table/col sep=comma},
        legend style={
            at={(0.5,0.94)},
            anchor=south, 
            legend columns=4,
            font=\footnotesize,
            /tikz/every even column/.append style={column sep=0.5cm}
        }
    ]
            \foreach \n in {1,...,10} {
            \addplot+[boxplot, fill, draw=black, boxplot/draw position={\n-0.3}, fill=zero, forget plot] table[y index=\n] {\zerodatatransposed};
        }
        \addlegendimage{fill=zero, draw=black, area legend}
        \addlegendentry{Zero}
        
        \foreach \n in {1,...,10} {
            \addplot+[boxplot, fill, draw=black, boxplot/draw position={\n-0.1}, fill=simple, forget plot] table[y index=\n] {\simpledatatransposed};
        }
        \addlegendimage{fill=simple, draw=black, area legend}
        \addlegendentry{\ours}
        
        \foreach \n in {1,...,10} {
            \addplot+[boxplot, fill, draw=black, boxplot/draw position={\n+0.1}, fill=roi, forget plot] table[y index=\n] {\roidatatransposed};
        }
        \addlegendimage{fill=roi, draw=black, area legend}
        \addlegendentry{\ourm}
        
        \foreach \n in {1,...,10} {
            \addplot+[boxplot, fill, draw=black, boxplot/draw position={\n+0.3}, fill=combo, forget plot] table[y index=\n] {\combodatatransposed};
        }
        \addlegendimage{fill=combo, draw=black, area legend}
        \addlegendentry{\ourl}
    \end{axis}
\end{tikzpicture}
        \caption{EPE$_\text{A}$}
    \end{subfigure}

    \caption{\textbf{Detailed evaluations} of our architectures for each action on \birdviewplus in the \textit{time split}. The action \textit{all} is the combined evaluation of all sequences independently of the action captured. Note that by construction, the evaluation of our \textit{zero} baseline always has an endpoint-error of $0$ when only considering the static pixels (EPE$_\text{S}$). Similarly, an evaluation of EPE$_\text{M}$ for our two special actions off-area and void is not possible, as there are never any moving pixels for these actions in any frame.}
    \label{fig:app:action-box-results}
\end{figure}

\cref{fig:app:action-box-results} shows a detailed evaluation for each action type of our proposed architectures. Overall, these evaluations show a clear trend: \ourl is our most accurate model across almost all actions and metrics, and \ourm is the second-best. The differences in the EPE of our models for the action \textit{together} compared to simpler actions, such as \textit{fast-walk}, also confirm our findings in \cref{sec:limit} and \cref{fig:qualitative-results}.

Notably, both of our mask-based architectures \ourm and \ourl handle the special actions \textit{off-area} and \textit{void} almost without any error, whereas \ours sometimes predicts motions in these frames. This shows that masking strategies eliminate most of the background noise produced by a simple optical flow estimator in our setting.

    \begin{figure}[ht]
        \centering
        \renewcommand{\tabularxcolumn}[1]{m{#1}}
        \noindent
        \begin{tabularx}{\linewidth}{@{} >{\centering\arraybackslash}m{1em} @{} X @{}}
            \rotatebox{90}{\scriptsize Input} & 
            \includegraphics[width=.95\linewidth]{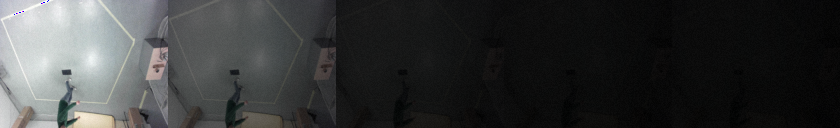} \\ [1ex] 
            
            \rotatebox{90}{\scriptsize MS-RAFT} & 
            \includegraphics[width=.95\linewidth]{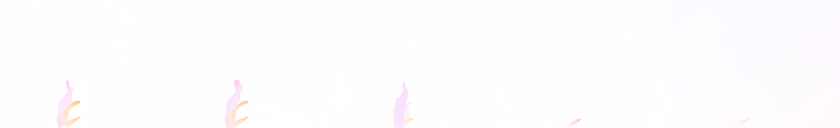} \\ 
            
            \rotatebox{90}{\scriptsize WiFlow} & 
            \includegraphics[width=.95\linewidth]{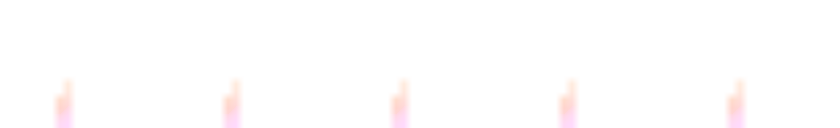}
        \end{tabularx}
        \caption{\textbf{Brightness dependency} of MS-RAFT and our \ourl. We gradually decrease the brightness and add additive noise to the original frame. Accuracy of MS-RAFT decreases while \ourl is not affected, as it is independent of the input images.}
        \label{tab:brightness}
    \end{figure}
    
\section{Optical Flow in Low Brightness Settings}
\label{supp:brightness}
Optical flow methods often depend heavily on brightness consistency. In \cref{tab:brightness}, we illustrate the brightness dependence of a state-of-the-art flow estimator (MS-RAFT~\cite{Jahedi:2024:HRM}) by simulating low-brightness settings. We decrease the brightness and add $3\%$ noise to simulate frames taken in darker environments. This results in a reduced accuracy compared to our method. %
     Further, there is ongoing research on brightness and other optical adversarial attacks~\cite{Zheng:2020:OFD,Schmalfuss:2025:BRI}, to which a CSI-based flow is invariant by design.

\FloatBarrier
{
    
}

\newpage

\end{document}